\documentclass{article} %
\usepackage{configuration/iclr2025_conference}
\usepackage{times}

\usepackage{amsmath,amsfonts,bm}

\def\eqref#1{equation~\ref{#1}}

\def\1{\bm{1}}

\def\vv{{\bm{v}}}

\def\mA{{\bm{A}}}

\def\mG{{\bm{G}}}

\def\mK{{\bm{K}}}

\def\mM{{\bm{M}}}

\def\mQ{{\bm{Q}}}

\def\mV{{\bm{V}}}

\DeclareMathAlphabet{\mathsfit}{\encodingdefault}{\sfdefault}{m}{sl}
\SetMathAlphabet{\mathsfit}{bold}{\encodingdefault}{\sfdefault}{bx}{n}

\newcommand{\R}{\mathbb{R}}

\usepackage{hyperref}
\usepackage{url}
\usepackage{amsmath,amsthm,amssymb}
\newtheorem{theorem}{Theorem}[section]

\newtheorem{remark}[theorem]{Remark}

\providecommand{\R}{\mathbb{R}} %

\providecommand{\1}{\mathbf{1}}

\providecommand{\kk}{\mathbf{k}}

\providecommand{\qq}{\mathbf{q}}

\providecommand{\vv}{\mathbf{v}}

\providecommand{\mA}{\mathbf{A}}

\providecommand{\mG}{\mathbf{G}}

\providecommand{\mK}{\mathbf{K}}

\providecommand{\mM}{\mathbf{M}}

\providecommand{\mQ}{\mathbf{Q}}

\providecommand{\mV}{\mathbf{V}}

\providecommand{\cO}{\mathcal{O}}

\newenvironment{talign*}
{\let\displaystyle\textstyle\csname align*\endcsname}
{\endalign}

\usepackage[utf8]{inputenc}         %
\usepackage[T1]{fontenc}            %
\usepackage{url}                    %
\usepackage{booktabs}               %
\usepackage{amsfonts}               %
\usepackage{nicefrac}               %
\usepackage{microtype}              %
\usepackage{xcolor}                 %

\usepackage[flushleft]{threeparttable}
\usepackage{float}
\usepackage{multirow}
\usepackage{xspace}
\usepackage{natbib}
\usepackage{enumitem}
\usepackage[font=small]{caption}
\usepackage{autobreak}
\usepackage{sidecap}
\usepackage{wrapfig}
\usepackage{bbding}
\usepackage[toc, page, header]{appendix}
\usepackage{pifont}

\usepackage[textwidth=3.5cm,textsize=tiny]{todonotes}
\newcommand{\algopt}{\textsc{DeFT}\xspace}
\newcommand{\flashdecoding}{Flash-Decoding\xspace}

\newcommand{\treeAttn}{Tree Attention-Medusa\xspace}
\newcommand{\kvcache}{KV cache\xspace}
\newcommand{\kvCache}{KV Cache\xspace}

\usepackage{subfig}
\usepackage{algorithm}
\usepackage{algorithmic}
\usepackage{tabularx}
\usepackage{makecell}
\title{DeFT: \underline{De}coding with \underline{F}lash \underline{T}ree-Attention for Efficient Tree-structured LLM Inference}

\author{%
    Jinwei Yao$^{1,4,}$\thanks{Equal contribution. Work done during Jinwei's visit to Westlake University.} \quad Kaiqi Chen$^{2,*}$ \quad Kexun Zhang$^{3}$  \quad Jiaxuan You $^{4,\dag}$ \\ \textbf{Binhang Yuan}$^{5}$ \quad \textbf{Zeke Wang}$^{2,\dag}$  \quad \textbf{Tao Lin}$^{1,}$\thanks{Corresponding author.} \\
\texttt{jinwei.yao1114@gmail.com}; \quad \texttt{\{chiaki\_cage,wangzeke\}@zju.edu.cn}; \\ \texttt{kexunz@andrew.cmu.edu}; \quad \texttt{jiaxuan@illinois.edu};\\ \texttt{biyuan@ust.hk}; \quad \texttt{lintao@westlake.edu.cn} \\
  $^1$Westlake University \quad
  $^2$Zhejiang University \quad
  $^3$Carnegie Mellon University \quad \\
  $^4$University of Illinois Urbana-Champaign \quad
  $^5$Hong Kong University of Science and Technology 
}

\iclrfinalcopy %
\begin{document}

\maketitle

\begin{abstract}
    Large language models (LLMs) are increasingly employed for complex tasks that process multiple generation calls in a tree structure with shared prefixes of tokens, including few-shot prompting, multi-step reasoning, speculative decoding, etc.
    However, existing inference systems for tree-based applications are inefficient due to improper partitioning of queries and KV cache during attention calculation.
    This leads to two main issues: (1) a lack of memory access (IO) reuse for KV cache of shared prefixes, and (2) poor load balancing.
    As a result, there is redundant KV cache IO between GPU global memory and shared memory, along with low GPU utilization.
    To address these challenges, we propose \algopt\footnote{By default, \algopt refs to \algopt-Flatten, which has \textbf{Flattened Tree KV Splitting} before loading KV cache for attention calculation. } (\underline{De}coding with \underline{F}lash \underline{T}ree-Attention), a hardware-efficient attention algorithm with prefix-aware and load-balanced KV cache partitions.
    \algopt reduces the number of read/write operations of KV cache during attention calculation through \emph{KV-Guided Grouping}, a method that avoids repeatedly loading KV cache of shared prefixes in attention computation.
    Additionally, we propose \emph{Flattened Tree KV Splitting}, a mechanism that ensures even distribution of the KV cache across partitions with little computation redundancy, enhancing GPU utilization during attention computations.
    By reducing 73-99$\%$ \kvcache IO and nearly 100$\%$ IO for partial results during attention calculation, \algopt achieves up to 2.23/3.59$\times$ speedup in the decoding/attention latency across three practical tree-based workloads compared to state-of-the-art attention algorithms. Our code is available at \url{https://github.com/LINs-lab/DeFT}.

    \looseness=-1
\end{abstract}

\setlength{\parskip}{1.25pt plus1.25pt minus0pt}

\section{Introduction}

\begin{wrapfigure}[31]{l}{0.48\textwidth}
    \centering
    \vspace{-1em}
    \includegraphics[width=0.45\textwidth]{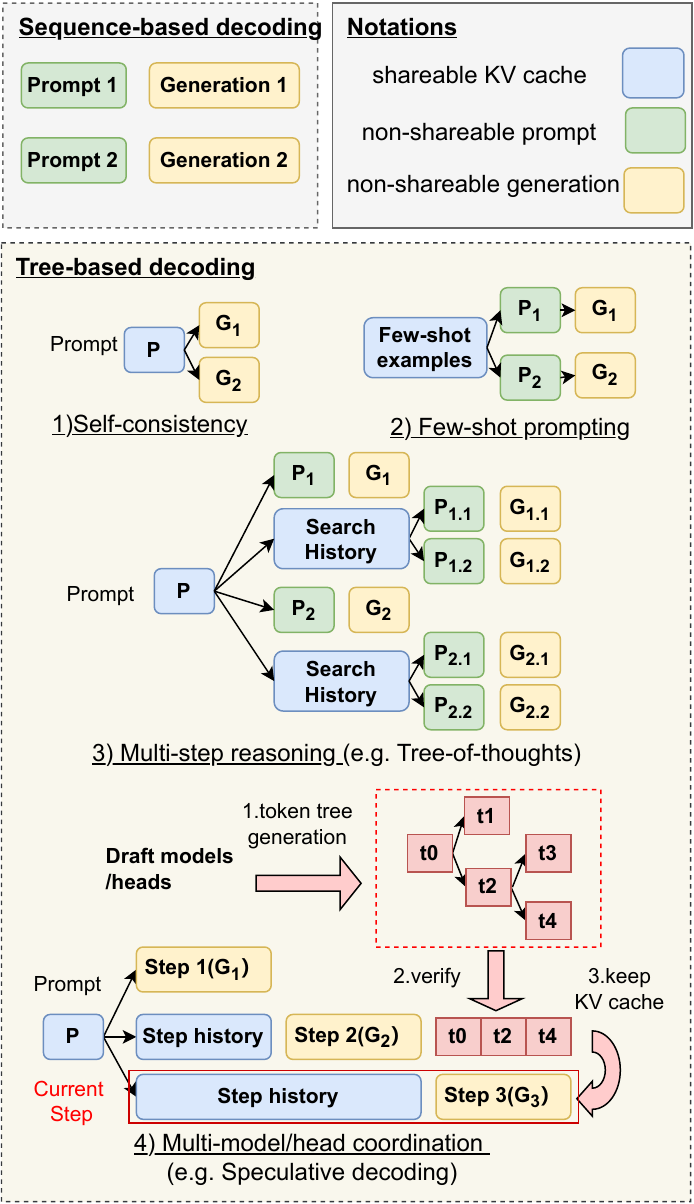}
    \caption{\textbf{An illustration of Sequence-based decoding and Tree-based decoding}.
    }
    \label{fig:treeVSseq}
\end{wrapfigure}

Large language models (LLMs) \citep{achiam2023gpt,touvron2023llama,touvron2023llama2} are extensively utilized across a range of tasks like chatbot~\citep{roller2020recipes}, code generation~\citep{mark2021carr}, reasoning~\citep{yao2023tree,besta2023graph,ning2023skeleton}, etc.
Traditionally, the interactions between LLMs and application users are sequential: the user sends a new prompt after completion result of the previous prompt is received.
However, many applications are now designed to process sequences with an internal tree structure, including self-consistency~\citep{wang2022self}, few-shot prompting~\citep{mann2020language}, multi-step reasoning~\citep{yao2023tree,hao2023reasoning,xie2024self}, and speculative decoding~\citep{miao2023specinfer,cai2024medusa}, etc, as shown in Figure~\ref{fig:treeVSseq}.
Usually, these applications produce substantially more tokens than traditional ones, to provide large space for tree search~\citep{graves2012sequence,lu2022neurologic,liu2023making} or selection, as shown in \autoref{tab:CotVSTot}.
\textbf{We need a more efficient decoding algorithm in response to this interaction paradigm change from sequence-based decoding to tree-based decoding}.

\begin{wraptable}[17]{r}{0.5\textwidth}
    \caption{\small \textbf{Comparison of efficiency in sequence-based CoT~\citep{wei2022chain} and tree-based ToT~\citep{yao2023tree} decoding for a reasoning task.}
        The task is \textit{sorting $128$ numbers} from \cite{besta2023graph}.
        The total generated tokens of CoT is only $525$ while $38{,}315$ in ToT, resulting in inefficiency in end-to-end latency (\texttt{second}) and IO (\texttt{TB}).
        IO mainly consists of two parts as follows.
        (i) \textit{KV cache}: \texttt{IO-KV};
        (ii) \textit{Partial results during attention calculation like $QK^T$ and softmax}: \texttt{IO-PA};
        Baselines:
        (i) \textit{\flashdecoding}~\citep{dao2023flashdecoding};
        (ii) \textit{Tree Attention}: tree attention in Medusa~\citep{cai2024medusa}.
        \label{tab:CotVSTot}
    }
    \vspace{-0.5em}
    \begin{huge}
        \resizebox{0.5\textwidth}{!}{ \renewcommand{\arraystretch}{1.0}
            \centering
            \vspace{-1em}
            \Huge
            \begin{tabular}{cccccc}
                \toprule
                                            & \texttt{Latency}              & \texttt{IO-KV} & \texttt{IO-PA } \\ \midrule
                \flashdecoding+ CoT         & 21                            & 0.6            & 0               \\
                \midrule
                \flashdecoding+ ToT         & 429.65                        & 59.96          & 0               \\
                Tree Attention + ToT        & 380.87                        & 12.40          & 3.69            \\
                DeFT-Flatten(ours) + ToT    & \textcolor{red}{94.67}        & 12.40          & 0               \\
                Speed up over best baseline & \textcolor{red}{$4.02\times$} & -              & -               \\
                \bottomrule
            \end{tabular} }

    \end{huge}
    \vspace{-1em}
\end{wraptable}

When requests have shared prefixes in a tree structure, existing inference systems~\citep{huggingface_text_generation_inference,nvidia_tensorrt_llm,kwon2023efficient} designed for sequence-based decoding introduce redundancy by failing to be prefix-aware at one or more of the following three levels:
(1) \emph{computation}---for instance, the redundant recomputation of KV caches for shared prompts across requests in a batch~\citep{huggingface_text_generation_inference};
(2) \emph{memory storage}---for example, the redundant storage of KV caches for shared prefixes~\citep{huggingface_text_generation_inference,kwon2023efficient,nvidia_tensorrt_llm};
(3) \emph{memory access (IO)}---such as repeatedly loading the KV cache of a shared system prompt during attention calculations~\citep{huggingface_text_generation_inference, kwon2023efficient,nvidia_tensorrt_llm}.
Although some tree-based inference systems~\citep{zheng2023efficiently,gim2023prompt,cai2024medusa,miao2023specinfer} address the first two issues, they largely overlook the third and arguably the most crucial aspect: \emph{memory access}, which is critical in the context of memory-bound LLM inference~\citep{shazeer2019fast,cai2024medusa,kim2023squeezellm}.

To accelerate the tree-structured LLM inference, an important question is whether we can leverage the shared patterns in multi-cascaded prefixes to design a faster and more memory-efficient attention algorithm.
This task is challenging due to two key issues as follows.
\textbf{C1: How to ensure prefix-awareness in memory access of KV cache?} Current memory-efficient attention algorithms~\citep{dao2022flashattention,dao2023flashdecoding,hong2023flashdecoding} are optimized for sequence-based decoding, which leads to a lack of prefix-awareness during memory access.
As a result, shared prefixes in the KV cache are repeatedly loaded.
\textbf{C2: How to split the tree-structured KV cache for load balancing and high GPU utilization?}
For optimal GPU utilization, the current KV splitting strategy for sequence-based decoding—Flash-Decoding~\citep{dao2023flashdecoding}, which splits sequence KV into chunks—cannot be directly applied to tree-structured KV.
Tree-structured KV caches also need to be effectively partitioned: however, if we naively split them by nodes, token lengths across different nodes can vary significantly (e.g., in speculative decoding~\citep{cai2024medusa}, some nodes might only have $1$ token while the root node could have thousands), making it difficult to maintain load balance and efficient computation.
\looseness=-1

To address the above challenges, we propose \algopt-Flatten, a prefix-aware tree attention algorithm with a flattened tree KV splitting strategy, based on two key insights.
$\bullet$ First, how queries and KV caches are grouped for attention calculation significantly impacts memory access.
Existing approaches use a \emph{\textbf{Q-Guided Grouping}} strategy, where each request/query is grouped with all corresponding KV caches.
While this eliminates IO redundancy for queries, the prefix KV cache still gets loaded multiple times.
To address \textbf{C1}, we propose \emph{\textbf{KV-Guided Grouping}}: \algopt-Flatten groups the prefix's KV cache with all shared queries, ensuring the prefix KV cache is only loaded once, significantly reducing redundant loading with negligible IO overhead for reloading queries.
The IO overhead for queries (Q) is minimal compared to the KV cache, as the maximum query length typically corresponds to the number of root-to-leaf paths in the tree, making the queries relatively short (e.g., dozens of tokens) compared to the KV cache length in each node (e.g., hundreds or thousands of tokens).
$\bullet$ Second, since LLM inference is IO-bound, the attention overhead of each QKV group is primarily influenced by the IO of the KV cache.
Therefore, it is crucial to ensure that the KV lengths of different QKV groups are nearly balanced.
To address \textbf{C2}, we propose a \emph{\textbf{Flattened Tree KV Splitting}}, which enables balanced partitions by dividing the flattened tree KV into even chunks, using bit causal masks to capture causal relationships between queries and KV cache.

We summarize our contributions as follows: \looseness=-1
\begin{itemize}[nosep, leftmargin=12pt]
    \item We propose a hardware-efficient tree attention algorithm---\algopt-Flatten, which is IO-aware of shared prefixes' KV cache and load-balanced in computation.
          \looseness=-1
    \item We implement \algopt-Flatten on OpenAI Triton \citep{tillet2019triton} to gain precise management over memory access and fuse all attention operations into a single GPU kernel.
    \item We theoretically justify the superiority of \algopt-Flatten over the existing attention algorithms \citep{wolf2019huggingface,dao2023flashdecoding,cai2024medusa,miao2023specinfer} in terms of IO complexity.
    \item We empirically verify its effectiveness on few-shot prompting, multi-step reasoning, and speculative-decoding tasks.
          \algopt-Flatten can achieve a decoding latency speedup of \textbf{1.3}$\times$ for few-shot prompting, \textbf{2.2}$\times$ for speculative decoding, \textbf{1.1}$\times$ for multi-step reasoning, due to an up to \textbf{3.59}$\times$ faster attention calculation, with the baseline implementations~\citep{dao2023flashdecoding,cai2024medusa,zheng2023efficiently}.
    \item We compare different tree split strategies---\algopt-Node, \algopt-Node-Chunk, and \algopt-Flatten in ablation studies (see section \ref{subsec:ablation}), showing the balanced partitioning of QKV groups matters.
\end{itemize}

\section{Related Work}
\textbf{Tree-based Decoding.}
Tree-based decoding, exemplified by beam search~\citep{graves2012sequence}, has been pivotal in NLP, handling lexical and logical constraints \citep{anderson-etal-2017-guided, post-vilar-2018-fast, hokamp-liu-2017-lexically}, mitigating gender bias \citep{lu2021neurologic}, achieving communicative goals \citep{holtzman2018learning}, and improving alignment \citep{liu2023making}.
Based on the structure feature of queries and KV cache, we can classify tree-based decoding into two patterns:
(i) Tree-structured past KV with parallel queries---usually in multi-step reasoning \citep{yao2023tree, besta2023graph, ning2023skeleton}, using search trees with parallel hypothesis generation and selection based on scoring functions, either score candidates per token \citep{dathathri2019plug, lu2021neurologic, lu2022neurologic} or per reasoning step \citep{welleck2022naturalprover, uesato2022solving,xie2024self}.
(ii) Past KV in sequence with tree-structured queries---usually in speculative decoding~\citep{cai2024medusa,miao2023specinfer}.
Further details on these two patterns are discussed in Appendix \ref{Tree_decoding_disc}.
Although tree-based search algorithms like A* \citep{lu2022neurologic} and Monte-Carlo Tree Search \citep{liu2023making} have been applied, the efficiency of tree-based decoding remains largely under-explored.
\looseness=-1

\textbf{Memory-efficient Attention Algorithms.}
Existing memory-efficient attention algorithms target sequence-based decoding.
FlashAttention~\citep{dao2022flashattention} improves self-attention computation in LLM training via tiling and kernel fusion, reducing IOs.
\flashdecoding \citep{dao2023flashdecoding} extends this, enhancing parallelism by dividing K and V and introducing global reduction to gather partial attention results, enabling efficient decoding for long sequences.
Unfortunately, applying these memory-efficient algorithms to the tree-based decoding overlooks redundancy in IO of tree-structured \kvcache, which is the focus of \algopt.
\looseness=-1

\textbf{Tree Attention.}
Integrated into LLM inference, tree attention reduces computation, storage, and kernel launching overheads~\citep{miao2023specinfer}.
Tree-structured token candidates undergo parallel decoding, with SpecInfer~\citep{miao2023specinfer} introducing a topology-aware causal masked tree attention algorithm, dynamically updating a causal mask to capture relationships among tokens.
Medusa~\citep{cai2024medusa} uses a similar mechanism with a static causal mask, while other works~\citep{ zhao2023lookahead,liu2024apar} adopt analogous approaches to enhance attention calculation efficiency.
However, unlike \algopt, these existing works utilizing tree attention do not take memory access into consideration.
\looseness=-1

\textbf{Storage Optimization of Tree-based Decoding.}
LLM frameworks optimized for tree-based decoding \citep{kwon2023efficient, zheng2023efficiently} focus on memory storage efficiency.
vLLM~\citep{kwon2023efficient} enhances GPU memory utilization, allowing sequences from the same parent to share \kvcache storage. SGLang~\citep{zheng2023efficiently} supports dynamic \kvcache management during multi-round interactions with LLMs, improving memory efficiency.

\textbf{Discussion on Concurrent Works.}
Some concurrent works~\citep{ye2024chunkattention,juravsky2024hydragen,athiwaratkun2024bifurcated,cascadeinference,zhu2024relayattention} also recognize the importance of IO during LLM inference.
However, these works have at least one of these flaws:
i) they~\citep{ye2024chunkattention,juravsky2024hydragen,athiwaratkun2024bifurcated,cascadeinference,zhu2024relayattention} cannot be easily extended to situations where the decoding tree has more than two levels---they target single-context batch sampling scenarios, a special case of general tree-based decoding with a system prompt as prefix and unique suffixes in the first depth;
ii) they~\citep{juravsky2024hydragen,athiwaratkun2024bifurcated} do not consider the inefficiency caused by the lengths of different nodes in the decoding tree.
See the comparison of \algopt and concurrent works in Appendix \ref{Concurrent_disc}.

\section{DeFT}
In this section, we first introduce the background knowledge of LLM inference, upon which we outline the importance of QKV partitions for attention calculation.
We then present the overview of \algopt algorithm and Attention Kernel design, with its system support.
Finally, we propose efficient QKV partitioning method for \algopt, which not only reduces memory access of prefixes' KV cache and partial results (e.g., Softmax), but also ensures balanced partitions during attention computation.

\subsection{Preliminary\label{subsec:Preliminary}}
\paragraph{LLM inference and its bottleneck.}
LLM inference involves two stages: (1) prefill and (2) decoding.
During the prefill stage, a prompt is tokenized to initialize LLM.
The output of the prefill stage becomes the input for the decoding stage.
The decoding stage is auto-regressive, with each output token from the previous step serving as the input token for the next step.
Due to the sequential process of auto-regressive decoding, LLM inference is memory-bound \citep{shazeer2019fast,kim2023squeezellm,cai2024medusa}, wherein every forward pass requires transferring all model parameters and \kvcache from slower but larger High-Bandwidth Memory (HBM) to the faster but much smaller shared memory of the GPU \citep{jia2021dissecting} \footnote{
    A100's HBM has 1.5-2TB/s bandwidth and 40-80GB;
    its shared memory has 19TB/s bandwidth and 20MB.
    \looseness=-1
}.
Another potential bottleneck is low GPU utilization~\citep{dao2023flashdecoding}, which happens when the parallelism (usually limited by the batch size is much smaller than the number of streaming multiprocessors (SMs) on the GPU ($108$ for an A100), where the operation will only utilize a small portion of the GPU.

\paragraph{The execution pattern of attention algorithms on GPUs.}
We can separate the execution of attention algorithms into two main phases:
(1) \textsc{QKV Preparation Phase}: group Query, Key, and Value (QKV) logically to partitions and map QKV groups to different streaming multiprocessors (SMs) of GPUs;
(2) \textsc{Attention Calculation Phase}: load QKV partitions to different SMs' shared memory and apply attention algorithms to each group for final attention results.
\looseness=-1

\paragraph{QKV partitions with segmented attention.}
In sequence-based decoding, QKV partitioning is crucial when the parallelism (usually limited by the batch size~\citep{dao2023flashdecoding}) is much smaller than the number of streaming multiprocessors (SMs) on the GPU ($108$ for an A100), where the operation will only utilize a small portion of the GPU.
To enable high GPU utilization, Flash-Decoding~\citep{dao2023flashdecoding} partitions the queries and KV cache then calculates the attention in parallel. \\
Details are as follows:
(1) \textsc{QKV Preparation Phase:} for each query in the batch, split its sequential KV cache into chunks as different QKV partitions.
(2) \textsc{Attention Calculation Phase:} it calculates segmented attention $A_0$, $A_1$, and $A_2$ over three segments, respectively, and then gets final attention by online Softmax merging~\citep{dao2022flashattention,dao2023flashdecoding} based on segmented attention from different QKV partitions.
We elaborate on the procedure below.
\looseness=-1
\begin{itemize}[nosep, leftmargin=12pt]
    \item Let's say we have key tensor $\mK \in \R^{l_{kv} \times d}$, value tensor $\mV \in \R^{ l_{kv} \times d }$, and query tensor $\mQ \in \R^{ l_q \times d }$.
          Considering the general case $\mK$ and $\mV$ are partitioned across the sequence (row) dimension into three parts for parallel calculation, respectively: $\mK = \mK_0 \parallel \mK_1 \parallel \mK_2$, and $\mV = \mV_0 \parallel \mV_1 \parallel \mV_2$, with ``$\parallel$'' denoting concatenation along the row axis.
    \item We calculate the attention $\mA_0$, $\mA_1$, and $\mA_2$ over KV chunks in different streaming-multiprocessors (SMs) of GPU, where
          $\mA_0 = \langle \mQ$, $\mK_0, \mV_0 \rangle$, $\mA_1 = \langle \mQ, \mK_1, \mV_1 \rangle$, $\mA_2 = \langle \mQ, \mK_2, \mV_2 \rangle$,
          and
          $
              \langle \qq, \kk, \vv \rangle = \operatorname{Softmax} \left(\nicefrac{\qq \kk^\top}{\sqrt{d}}\right) \vv \,.
          $
    \item We calculate $\text{LogSumExp}$ $(\operatorname{LSE})$ as a weight of merging $\mA_0$, $\mA_1$, and $\mA_2$.
          We define $ \operatorname{LSE}(\qq, \kk) = \log\left(\sum\left(\exp\left( \nicefrac{ \qq \kk^\top }{\sqrt{d}} \right)\right)\right) $.
    \item We have $\langle \mQ, \mK, \mV \rangle = \operatorname{SegAttn}(\mA_0, \mA_1, \mA_2)$, which means segmented attention with Online Softmax~\citep{dao2022flashattention}:
          \begin{equation}
              \begin{aligned}
                  \textstyle
                  \operatorname{SegAttn}(\mA_0, \mA_1, \mA_2)= \frac{
                      \mA_0 e^{\operatorname{LSE}(\mQ, \mK_0)} + \mA_1 e^{\operatorname{LSE}(\mQ, \mK_1)} + \mA_2 e^{\operatorname{LSE}(\mQ, \mK_2)}
                  }{
                      e^{\operatorname{LSE}(\mQ, \mK_0)} + e^{\operatorname{LSE}(\mQ, \mK_1)} + e^{\operatorname{LSE}(\mQ, \mK_2)}
                  } \,, \text{ where } e := \text{exp} \,.
              \end{aligned}
              \label{eq:seg_softmax}
          \end{equation}

\end{itemize}

\begin{figure}[!t]
    \centering
    \includegraphics[width=0.8\textwidth]{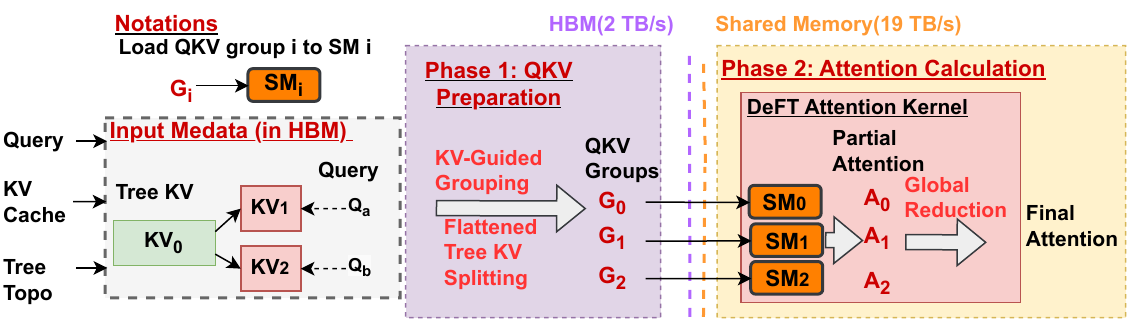}
    \vspace{-0.5em}
    \caption{
        \textbf{Overview of \algopt}.  \textit{Input Metadata
        } is prepared in the system elaborated in Appendix \ref{comp_func}. In \emph{QKV Preparation Phase} (see Section \ref{sec:DeFT_Attn}), the QKV will be grouped logically to partitions with IO-awareness of shared prefixes' KV cache and load-balancing. These partitions will guide the loading of QKV on the \emph{Attention Calculation Phase} (see Appendix \ref{A:Tech}), where the attention calculation will be executed.
    }
    \vspace{-0.5em}
    \label{fig:DeFT}
\end{figure}

\subsection{Overview of \algopt \label{subsec:sys_overview}}
\paragraph{The importance of QKV partitions.}
For tree-based decoding, logically partitioning QKV is necessary for attention calculation with high parallelism.
The branch number of tree-structured generation requests may be insufficient to fully utilize the GPU when the number of tokens in the tree-structured KV cache is large, due to memory capacity limitations.
For example, a request for the reasoning task of sorting $128$ numbers~\citep{besta2023graph}, involves around 40K tokens in a Llama2-7B model, whose KV cache occupies 20GB, which means an 80GB A100 can only  process at most $4$ requests with such token numbers.

\paragraph{Motivation of \algopt.}
\algopt aims to address two potential bottlenecks (i.e., IO and GPU utilization) of LLM inference when dealing with tree-structured KV sequences.
Let's say we have a simple tree with two cascades, as shown in the left part of~\autoref{fig:DeFT}: for two queries $\mQ_a$ and $\mQ_b$, the corresponding keys satisfy $\mK_a = \mK_0 \parallel \mK_1$ and $\mK_b = \mK_0 \parallel \mK_2$, respectively, and values obey the same rule.
\algopt is designed to:
(1) minimize IO by eliminating redundant memory access of the shared prefix's KV cache ($\mK_0$ and $\mV_0$) for $\mQ_a$ and $\mQ_b$;
(2) ensure balanced workloads for high GPU utilization, so that the overhead of computing each segmented attention $\mA_i$ remains nearly identical.
Since the global reduction in \eqref{eq:seg_softmax} requires all partial attention, if the overhead for computing $\mA_i$ is significantly larger than $\mA_j$, the SM responsible for calculating $\mA_j$ will experience prolonged idleness.
\looseness=-1

\paragraph{Technique overview of \algopt.}
\algopt aims to be a hardware-efficient attention algorithm by reducing memory access and ensuring load-balancing for tree-based decoding.
See details in \autoref{fig:DeFT}:
\looseness=-1
\begin{itemize}[nosep, leftmargin=12pt]
    \item[\ding{192}] In the \textsc{QKV Preparation Phase}, for prefix-aware and load-balanced QKV partitions, we introduce a \emph{KV-Guided Grouping} strategy to reuse the KV cache IO of the shared prefixes, and a \emph{Flattened Tree KV Splitting} for high GPU-utilization due to balanced and parallel attention calculation.
          See details in Section \ref{sec:DeFT_Attn}.
    \item[\ding{193}] During the \textsc{Attention Calculation Phase}, we design the \textsc{\algopt Attention Kernel}\footnote{GPUs utilize a vast array of threads to execute operations known as \textit{kernels}} to load QKV splits in a memory efficient way, which is logically grouped by the \textsc{QKV Preparation Phase}, then to perform the attention calculation.
          Key techniques are as follows, with details deferred in Appendix \ref{A:Tech}:
          1) Common \textit{Kernel Fusion} and \textit{Tiling} strategies avoid significant IO operations for partial results (i.e.. $\mQ \mK^\top$ and $\text{Softmax}$), which \treeAttn~\citep{cai2024medusa} lacks.
          2) \textsl{Tree-Topology-Aware Global Reduction}, which extends the global reduction mechanism from \flashdecoding~\citep{dao2023flashdecoding}.
          This approach efficiently computes the final attention for each query by aggregating partial attention results from QKV groups while considering the tree structure.
\end{itemize}

\paragraph{System frameworks of \algopt.}Apart from efficient \textsc{\algopt Attention Kernel}, our system for \algopt has other two advantages:
1) efficient memory management of the KV cache in a tree structure, and 2) flexible control of the tree decoding process with arbitrary user-defined functions to decide when and how to branch/prune.
The details of key components and their coordinations in the system refer to Appendix \ref{comp_func}.
\looseness=-1

\subsection{ Prefix-aware and Balanced Tree-structured \kvCache Partitions\label{sec:DeFT_Attn}}

\begin{figure}[!h]
    \begin{center}
        \includegraphics[width=1.\textwidth]{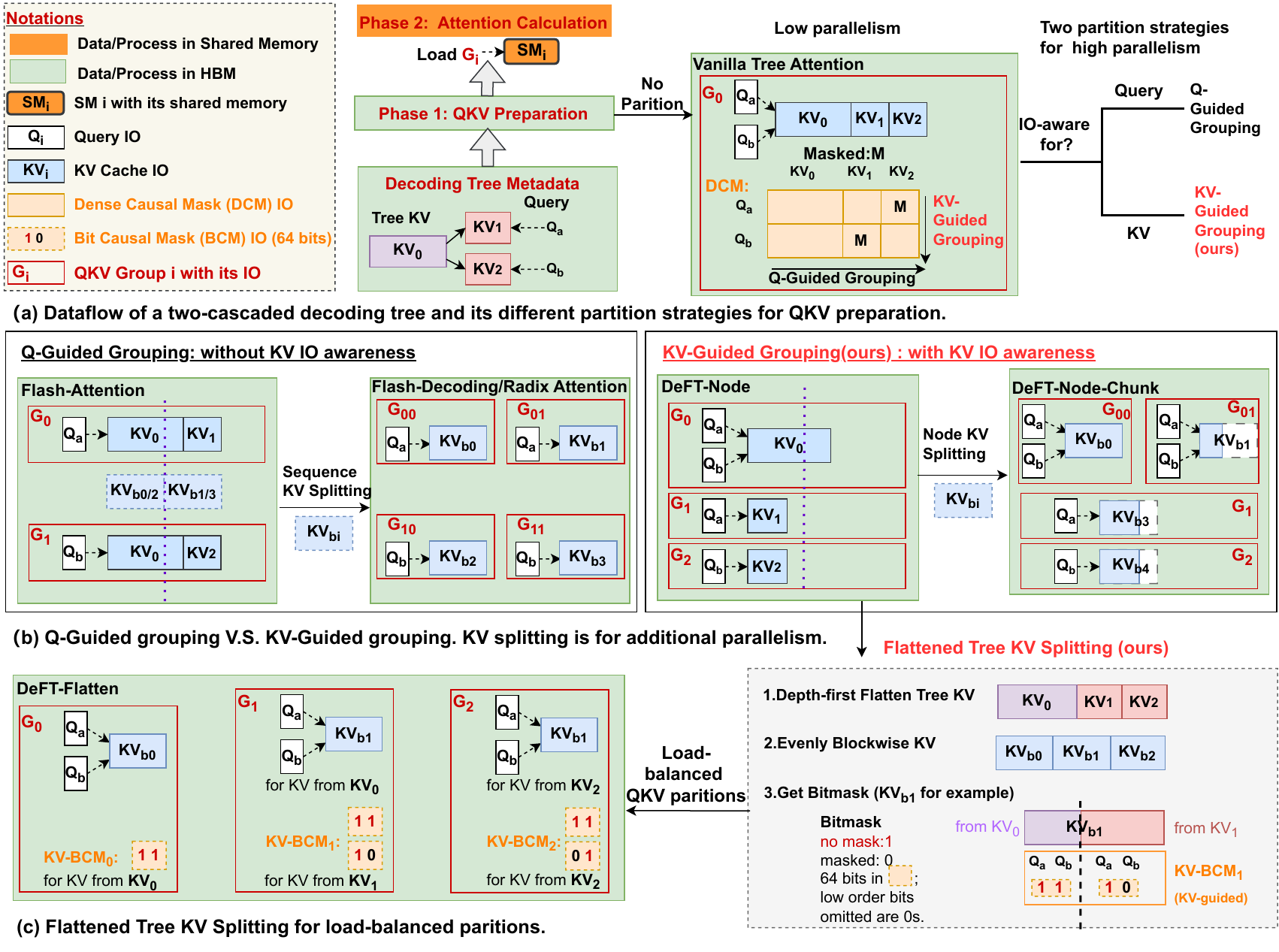}
    \end{center}
    \vspace{-1.2em}
    \caption{\small
        \textcolor{black}{\textbf{Comparison of QKV partitioning strategies during the QKV Preparation Phase between \algopt-Node/Node-Chunk/Flatten and different attention algorithm baselines.}} Note that the partitioning is logically designed without incurring any data movement costs for QKV. The amount of IO between the GPU HBM and shared memory required by each group is highlighted in \textcolor{red}{red rectangles}. \textcolor{black}{Part (a) illustrates the dataflow of a two-cascaded decoding tree example and three categories of QKV partitioning strategies: no partition(Vanilla Tree Attention), Q-Guided Grouping and KV-Guided Grouping.} The partitioning strategy will guide the loading of QKV during the subsequent \emph{Attention calculation phase}, where each QKV group $G_i$ will be loaded into $SM_i$ on the GPU. \textcolor{black}{Part (b) shows the comparison of Q-Guided Grouping and KV-Guided Grouping, where the latter can be IO-aware of prefix KV cache $KV_0$ and only load it once. \algopt-Node-Chunk is a weak load-balancing improvement of \algopt-Node by splitting large nodes (e.g., $KV_0$) to chunks.} \textcolor{black}{Part (c) illustrates the details (discussed in Remark \ref{rmk:flatten}) of Flattened Tree KV Splitting in \algopt-Flatten for load-balanced partitions, including Depth-first Flatten strategy, Evenly block-wise strategy, and Bit mask.}
        For a summary of baselines and \algopt, see \autoref{tab:grouping}. 
        \textcolor{black}{See analysis of tree-attention baselines~\citep{cai2024medusa,miao2023specinfer} in Remark \ref{rmk:group_treeattn}.}
        \label{fig:QKV_prepare}}
\end{figure}

This section delves into the details of the \textsc{QKV Preparation Phase}, which is a key design aspect of \algopt.
The discussion of the \textsc{Attention Calculation Phase} is deferred to Appendix \ref{A:Tech}.

Let's begin with a decoding example using the tree-structured KV cache shown in \autoref{fig:DeFT}.
If we group the entire tree-structured KV cache and queries into \(\mG_0\) without any partitions, we can refer to the Vanilla Tree Attention method illustrated in the \textcolor{black}{part (a) of}  \autoref{fig:QKV_prepare}.
This method calculates attention for all queries simultaneously in a single Streaming Multiprocessor (SM), with the aid of a dense causal mask (DCM).

However, this approach is inefficient due to low GPU utilization, as discussed in Section \ref{subsec:sys_overview}.
To address this inefficiency, effective partitioning of QKV is essential. This process involves two key considerations: (1) prefix awareness to minimize memory access to the KV cache and (2) load balancing to ensure even distribution of workloads across GPUs.

\paragraph{Q-Guided vs.\ KV-Guided Grouping.}
Most existing memory-efficient attention algorithms \citep{dao2022flashattention,dao2023flashdecoding,zheng2023efficiently} adopt \emph{Q-Guided Grouping} for QKV partitioning, where each query serves as the indicator for partitioning, grouping with its corresponding KV cache.
However, this method is not prefix-aware, e.g., in Flash-Attention (as shown in \autoref{fig:QKV_prepare}) $\mathbf{KV_0}$ is loaded twice, namely once for $\mQ_a$ and again for $\mQ_b$.

We resort to alternative \emph{KV-Guided Grouping} approach: by grouping each node's KV cache with all the queries that share it, the partitioning can be made prefix-aware, therefore reducing memory access to the KV cache.
For example, \algopt-Node (shown in \autoref{fig:QKV_prepare}) only loads the prefix KV cache $\mathbf{KV_0}$ once for attention computation.
The additional IO cost for queries is negligible since each query only contains a single token, while the KV cache may contain thousands of tokens.

\paragraph{Tree KV Splitting and Load-Balancing.}
Thanks to \emph{KV-Guided Grouping}, \algopt-Node is prefix-aware for KV cache IO.
However, it introduces a potential bottleneck: unbalanced workloads across different SMs.
For example, as seen in \algopt-Node of \autoref{fig:QKV_prepare}, $\mathbf{KV_0}$ might contain $1{,}000$ tokens, while $\mathbf{KV_1}$ only contains $2$ tokens.
If $\mG_0$ and $\mG_1$ are assigned to $SM_0$ and $SM_1$ respectively, $SM_1$ completes computation much earlier and remains idle, leading to  low SM utilization\footnote{
    Considering a microbenchmark that \algopt-Node with 64 queries shares a prompt of 4k tokens, the SM utilization is below $5\%$ for $82.35\%$ time of attention computation, as shown in \autoref{tab:microbench_gpu}.
}.

To address this, we need to balance the QKV partitions more evenly.
A straightforward approach is to chunk $\mK_0$, $\mK_1$, and $\mK_2$ at the physical level, while maintaining node-wise partitioning at the logical level, as shown in \algopt-Node-Chunk from \autoref{fig:QKV_prepare}.
However, this load-balancing strategy is weak: it only breaks large nodes (e.g., prompts with around 1k tokens) into smaller KV chunks, and it does not handle cases with many small nodes (e.g., speculative decoding), which could slow down inference due to more rounds of GPU execution for additional QKV groups.

\begin{table}[!t]
    \setlength\tabcolsep{2pt}
    \centering
    \small
    \caption{\small
        \textbf{Comparison of QKV partitioning strategies for baselines (most of which are shown in \autoref{fig:QKV_prepare}) and \algopt.}
        For IO redundancy, significant issues are highlighted in \textcolor{red}{red}, while negligible ones are in \textcolor{blue}{blue}.
        ``Q'' refers to queries, and ``KV'' refers to the KV cache.
        ``DCM'' stands for Dense Causal Mask (a matrix), and ``BCM'' refers to Bit Causal Mask (a set of 64-bit integers).
        ``PA'' represents partial results during attention calculations, including $\mQ \mK^T$, $\text{Softmax}$, etc.
        More $\star$ symbols indicate better-balanced workloads for QKV partitions.
        Details on IO complexity can be found in Appendix \ref{IOAna}.
        \label{tab:grouping}
    }
    \vspace{-1em}
    \begin{footnotesize}
        \resizebox{1.0\columnwidth}{!}{
            \begin{tabular}{@{}ccccccc@{}}
                \toprule
                \textbf{Attention Algorithm}                           & \textbf{Grouping Indicator} & \textbf{KV Split Granularity} & \textbf{IO Redundancy}                        & \textbf{Load-balancing Level} \\
                \midrule
                \textbf{Flash-Attention~\citep{dao2022flashattention}} & Q-guided                    & -                             & \textcolor{red}{KV}                           & $\star$                       \\
                \textbf{Flash-Decoding~\citep{dao2023flashdecoding}}   & Q-guided                    & by block                      & \textcolor{red}{KV}                           & $\star \star \star$           \\
                \textbf{Radix Attention~\citep{zheng2023efficiently}}  & Q-guided                    & by block                      & \textcolor{red}{KV}                           & $\star \star \star$           \\
                \textbf{Tree Attention-S~\citep{miao2023specinfer}}    & Q-guided                    & by block                      & \textcolor{red}{KV} and \textcolor{blue}{BCM} & $\star \star \star$           \\
                \textbf{Tree Attention-M~\citep{cai2024medusa}}        & entire tree                 & by GEMM in PyTorch            & \textcolor{red}{DCM} and \textcolor{red}{PA}  & $\star \star \star$           \\
                \textbf{Vanilla Tree Attention}                        & entire tree                 & no split                      & \textcolor{red}{DCM} and \textcolor{red}{PA}  & $\star$                       \\
                \textbf{\algopt-Node}                                  & KV-guided                   & by tree node                  & \textcolor{blue}{Q}                           & $\star$                       \\
                \textbf{\algopt-Node-Chunk}                            & KV-guided                   & by tree node, then by block   & \textcolor{blue}{Q}                           & $\star \star$                 \\
                \textbf{\algopt-Flatten}                               & KV-guided                   & by block                      & \textcolor{blue}{Q} and \textcolor{blue}{BCM} & $\star \star \star$           \\
                \bottomrule
            \end{tabular}
        }
    \end{footnotesize}
    \vspace{-1.4em}
\end{table}

As KV cache loading is the primary bottleneck for attention computation~\citep{cai2024medusa,tang2024quest}, it is important to achieve an even token length in each KV cache partition.
Therefore, we propose \algopt-Flatten, elaborated in Remark \ref{rmk:flatten}.

\begin{remark}[Techniques of \emph{Flattened Tree KV Splitting}\label{rmk:flatten}]
    As shown in the \textcolor{black}{part (c)} of \autoref{fig:QKV_prepare}, there are three key components:
    \begin{itemize}[nosep, leftmargin=12pt]
        \item \underline{\emph{Depth-first Flatten strategy.}}
              This approach minimizes redundant query IO and computation by leveraging the hierarchical relationship between parent and child KV nodes-- for instance, queries for parent $\mathbf{KV_0}$ (e.g., $\mQ_a$ and $\mQ_b$) include those for child $\mathbf{KV_1}$ (e.g., $\mQ_a$).
              Depth-first flattening instead of breadth-first, maximizes query overlap across KV cache from different nodes but allocated to the same chunk, reducing redundant computations like masked portions in $\mQ\mK^T$.
             
        \item \underline{\emph{Evenly block-wise strategy.}}
              It is the core of the splitting, where it ensures equal lengths of KV in each QKV group for balanced workloads of streaming multiprocessors (SMs) in GPUs.
        \item \underline{\emph{Bit mask~\citep{miao2023specinfer}.}}
              It is a set of $64$-bit integers used to record causal information of tokens in the tree.
              Therefore, its IO overhead (e.g., two 64-bit integers in KV-BC$M_1$ on \textcolor{black}{part (c)} of \autoref{fig:QKV_prepare}) is negligible compared to the dense causal mask~\citep{cai2024medusa}.
    \end{itemize}
\end{remark}

\begin{remark}[Discussion on Tree Attention Algorithms\label{rmk:group_treeattn}]
    Existing attention algorithms are designed for speculative decoding, where attention is calculated for the entire tree-structured queries.
    However, these methods are not memory-efficient.
    \textcolor{black}{For partition details, see \autoref{fig:treeattn_IO} in Appendix \ref{IOAna}.}
    \begin{itemize}[nosep, leftmargin=12pt]
        \item \underline{\emph{Tree Attention-Medusa~\citep{cai2024medusa}.}}
              Based on Vanilla Tree Attention (shown on the left in \autoref{fig:QKV_prepare}), this method uses PyTorch's General Matrix Multiply (GEMM) to partition Q and KV tensors.
              It is memory-inefficient for two reasons:
              (1) it does not utilize Flash-Attention to reduce memory access during the computation of intermediate results (e.g., Softmax);
              (2) it introduces a dense causal mask, whose memory access is significant.
        \item \underline{\emph{Tree Attention-SpecInfer~\citep{miao2023specinfer}.}}
              This algorithm employs \emph{Q-Guided Grouping} based on Vanilla Tree Attention and partitions the KV sequence through Flash-Decoding.
              It is memory-inefficient in redundantly loading the entire tree-structured KV cache for each query.
    \end{itemize}
\end{remark}

\paragraph{IO complexity analysis.}
We show \algopt-Flatten is better than existing attention algorithms in IO complexity, including Flash-Decoding~\citep{dao2023flashdecoding}, Tree Attention-Medusa~\citep{cai2024medusa}, and Tree Attention-SpecInfer~\citep{miao2023specinfer}.
See Appendix \ref{IOAna}.

\paragraph{Implementation details.}
We implement the \algopt attention kernel by OpenAI Triton \citep{tillet2019triton}, which enables us to control memory access from global memory to shared memory and attention calculations in a thread block granularity.
\algopt-Node and \algopt-Flatten algorithms with two phases in a Python style can be found in Appendix \ref{A:DeFT-alg} and Appendix \ref{A:DeFT-Sub-alg}, respectively.

\section{Experiments\label{Sec:exp}}
In this section, to demonstrate the effectiveness of \algopt under different tree topologies, we comprehensively conduct experiments on three types of tree-based decoding tasks, including:
(1) few-shot prompting~\citep{mann2020language}: a typical case study of tree-structured interactions with two levels--a prefix and several suffixes;
(2) multi-step reasoning~\citep{yao2023tree,xie2024self,hao2023reasoning}: tasks characterized by tree-structured past KV with parallel queries;
(3) speculative decoding~\citep{cai2024medusa,miao2023specinfer}: tasks involving past KV in sequence with tree-structured queries.
\looseness=-1

\subsection{Experimental Setup} \label{subsec:exp_setup}
\paragraph{Baselines.}

\begin{table}[!t]
    \setlength\tabcolsep{2pt}
    \centering
    \small
    \caption{\small \textbf{Comparison of baselines and \algopt.}
        Attention kernels of baselines are implemented to fit its memory management.
        Therefore, for a fair comparison with baselines, we implement \algopt-Node and \algopt-Flatten that fit both paged~\citep{kwon2023efficient}/unpaged memory management.
        \label{tab:baseline}
    }
    \vspace{-1em}
    \begin{footnotesize}
        \resizebox{1.0\columnwidth}{!}{
            \begin{tabular}{@{}ccccc@{}}
                \toprule
                \textbf{Method}         & Flash-Decoding~ & Tree Attention-Medusa & Radix Attention\citep{zheng2023efficiently} & \algopt       \\
                \midrule
                \textbf{Memory}         & unpaged         & unpaged               & paged                                       & unpaged/paged \\
                \textbf{Implementation} & Triton          & PyTorch               & Triton                                      & Triton        \\
                \bottomrule
            \end{tabular}
        }
    \end{footnotesize}
    \vspace{-1em}
\end{table}

\begin{table}[!t]
    \setlength\tabcolsep{2pt}
    \centering
    \small
    \caption{\small
        \textbf{Workloads generation}.
        ToT-BFS stands for Tree-of-Thoughts~\citep{yao2023tree} using breadth-first search.
        APPS~\citep{hendrycks2021measuring} is a competitive programming problem dataset.
        Medusa~\citep{cai2024medusa} is a speculative decoding framework.
        ``GoT'' stands for Graph-of-Thoughts~\citep{besta2023graph}, which contains iteration records using GPT-3.5 for complex reasoning tasks within ToT-BFS.
        See more details in \autoref{tab:workloadtree}.
        \label{tab:workload}
    }
    \vspace{-1em}
    \begin{footnotesize}
        \resizebox{1.0\columnwidth}{!}{
            \begin{tabular}{@{}ccccc@{}}
                \toprule
                \textbf{Task}        & \textbf{Prompt Dataset} & \textbf{Decoding Tree Source} & \textbf{Decoding Tree Collection Method}                   & \textbf{Stopping Criteria}         \\
                \midrule
                Few-shot prompting   & APPS                    & -                             & Pad the prompt to 4000 tokens                                                          & 400 iterations                     \\
                Multi-step reasoning & 4 tasks in GoT          & ToT-BFS                       & Reconstruct from interaction records with GPT 3.5 in GoT   & End of task($\sim$ 3500 iterations)                        \\
                Speculative decoding & APPS                    & Medusa                        & Record token tree shape and accepted token length per step &  $\sim$1000 steps(max length=6000) \\
                \bottomrule
            \end{tabular}
        }
    \end{footnotesize}
    \vspace{-1.5em}
\end{table}

We evaluate the performance of \algopt in NVIDIA A100 (80GB) in Llama3-8B model~\citep{touvron2023llama2} with the SOTA attention algorithms in sequence-based and tree-based decoding, as shown in \autoref{tab:baseline}.
Note that we did not include the tree attention operator of SpecInfer~\citep{miao2023specinfer} to our baselines as its kernel only supports at most 64 tokens in the token tree (the decoding tree except for the past sequence KV part), which is unsuitable for tree-based decoding with tree-structured KV (c.f.\ details in Appendix \ref{Tree_decoding_disc}).

\paragraph{Workloads generation.}
To ensure fairness for workloads of different baselines, we reconstruct decoding trees from real multi-step reasoning and speculative decoding tasks, as shown in \autoref{tab:workload}.
For multi-step reasoning, we include these four tasks from~\cite{besta2023graph}: (1) Sorting 128 numbers (\textit{Sorting} in short), (2) Document merging (\textit{Document} in short), (3) Keyword counting (\textit{Keyword} in short), and (4) Set intersection (\textit{Set} in short).
The tree decoding process would be forced to branch and prune the tree in certain iterations to get the same shape of the decoding tree as the original decoding tree sources.
See workload generation details and analysis in Appendix \ref{A:exp}.
\looseness=-1

\begin{table}[!t]
    \setlength\tabcolsep{2pt}
    \centering
    \small
    \caption{\small
        \textbf{Comparison of \algopt-Flatten and baselines in average decoding latency (in seconds) for tree-based decoding.}
        Here, $b$ represents the tree width, and $t$ denotes the token tree size (i.e., the number of tree-structured queries).
        The fastest method is in \textbf{bold}, and the second fastest is \underline{underlined}. \textcolor{blue}{\textit{Radix Attention}} is the best baseline in decoding latency.
        $\star$ denotes out-of-memory (OOM) errors for the A100 80GB GPU.
        \textit{Speedup Upper-bound (no attention)} refers to the maximum speedup we could achieve for  \textcolor{blue}{\textit{Radix Attention}} if we exclude the attention computation and only run other components including MLP.
        For more details on attention speedup, see \autoref{tab:e2eAttn_comp}.
        \label{tab:e2e_comp}
    }
    \vspace{-1.5em}
    \begin{footnotesize}
        \resizebox{1.0\columnwidth}{!}{
            \begin{tabular}{ccccccccccccc}
                \toprule
                \multirow{2}{*}{Memory} & \multirow{2}{*}{Method}                                                & \multicolumn{3}{c}{Few-shot Prompting }
                                        & \multicolumn{4}{c}{Multi-Step Reasoning}
                                        & \multicolumn{4}{c}{Speculative Decoding}
                \\ \cmidrule(lr){3-5} \cmidrule(lr){6-9} \cmidrule(lr){10-13}
                                        &                                                                        & \texttt{b=20}                           & \texttt{b=30}      & \texttt{b=50}
                                        & \textit{Sorting}                                                       & \textit{Document}                       & \textit{Keyword}   & \textit{Set}
                                        & \texttt{t=32}
                                        & \texttt{t=64}                                                          & \texttt{t=128}                          & \texttt{t=256}
                \\ \midrule
                \multirow{2}{*}{Unpaged}
                                        & \flashdecoding                                                         & 78.96                                   & 131.19             & 191.09
                                        & 429.65                                                                 & 241.20                                  & 32.75              & 51.76
                                        & 574.50                                                                 & 1128.45                                 & $\star$            & $\star$

                \\
                                        & \treeAttn                                                              & 52.58                                   & 103.90             & 144.07
                                        & 380.87                                                                 & 236.86                                  & 33.52              & 50.10
                                        & 263.40                                                                 & 483.35                                  & 924.97             & 1881.51

                \\

                \midrule
                \multirow{3}{*}{Paged}
                                        & \textcolor{blue}{\textit{Radix Attention}}                             & \underline{12.37}                       & \underline{14.08}  & \underline{16.54}
                                        & \underline{104.79}                                                     & \underline{69.61}                       & \underline{11.25}  & \underline{17.03}
                                        & \underline{54.66}                                                      & \underline{69.75}                       & \underline{108.56} & \underline{188.66}
                \\
                                        & \algopt-Flatten                                                        & \textbf{9.98}                           & \textbf{10.99}     & \textbf{12.48}
                                        & \textbf{94.67}                                                         & \textbf{66.95}                          & \textbf{10.90}     & \textbf{16.10}
                                        & \textbf{42.23}                                                         & \textbf{46.60}                          & \textbf{56.96}     & \textbf{84.27}
                \\  \midrule
                                        & \textit{Attention Speedup over \textcolor{blue}{Radix Attention} } & 1.73$\times$                            & 1.63$\times$       & 1.70$\times$
                                        & 1.39$\times$                                                           & 1.15$\times$                            & 1.21$\times$       & 1.34$\times$
                                        & 1.96$\times$                                                 & $2.41\times$                            & $3.11\times$           & $3.59\times$
                \\
                \midrule
                                        & \textit{Decoding Speedup over \textcolor{blue}{Radix Attention} }           & 1.24$\times$                            & 1.28$\times$       & 1.33$\times$
                                        & 1.10$\times$                                                           & 1.03$\times$                            & 1.03$\times$       & 1.05$\times$
                                        & 1.29$\times$                                                           & $1.50\times$                            & $1.91\times$       & $2.23\times$
                \\   \midrule                      &\textit{Speedup Upper-bound(no attention)}           & 1.71$\times$                               & 2.08$\times$& 2.51$\times$
                                        & 1.96$\times$                                                           & 1.82$\times$                            & 1.70$\times$       & 1.76$\times$
                                        & 1.89$\times$                                                           & 2.89$\times$                            & 3.34$\times$       & 4.36$\times$

                \\

                \bottomrule
            \end{tabular}
        }

        \vspace{-1.5em}
    \end{footnotesize}
\end{table}

\subsection{Analysis of Memory Management and Bottleneck} \label{subsec:mem}
As shown in \autoref{tab:baseline}, the kernel implementations of different attention algorithms adapt to different memory management.
To fairly compare their performance of wall-clock time speedup, we need to analyze the influence of memory management and the bottleneck of the system.

\paragraph*{A trade-off between memory storage and memory operation.}
In tree-based decoding, storing the KV cache for each branch is simple but lacks shared storage for the prefix’s KV cache. Given the limited GPU memory, not accounting for the tree structure in KV sharing reduces the number of tokens the tree can handle. Although storing KV caches by each tree node significantly improves storage efficiency, most attention kernels are designed for sequence-based decoding~\citep{dao2022flashattention,hong2023flashdecoding,dao2023flashdecoding}. To use these kernels, KV caches from different nodes must be concatenated into a single tensor, leading to substantial data movement costs~\citep{kwon2023efficient}.

\paragraph*{The benefits of paged memory for tree-based decoding.}
For efficient KV cache memory management, paged memory~\citep{kwon2023efficient,zheng2023efficiently} is the current mainstream technology.
These KV cache tensors are stored in a non-contiguous, paged layout to provide token-level reuse.
Besides higher storage efficiency, we note an additional benefit of paged memory management for tree-based decoding: non-contiguous storage in a memory pool is addressed by pointers, ensuring no need to materialize the tree-structured KV into a single tensor before executing the attention kernel.
Instead, we only need to record the memory pool addresses of each token's KV cache.

\begin{wrapfigure}{r}{0.4\textwidth}
    \vspace{-0.8cm}
    \centering
    \includegraphics[width=0.85\linewidth]{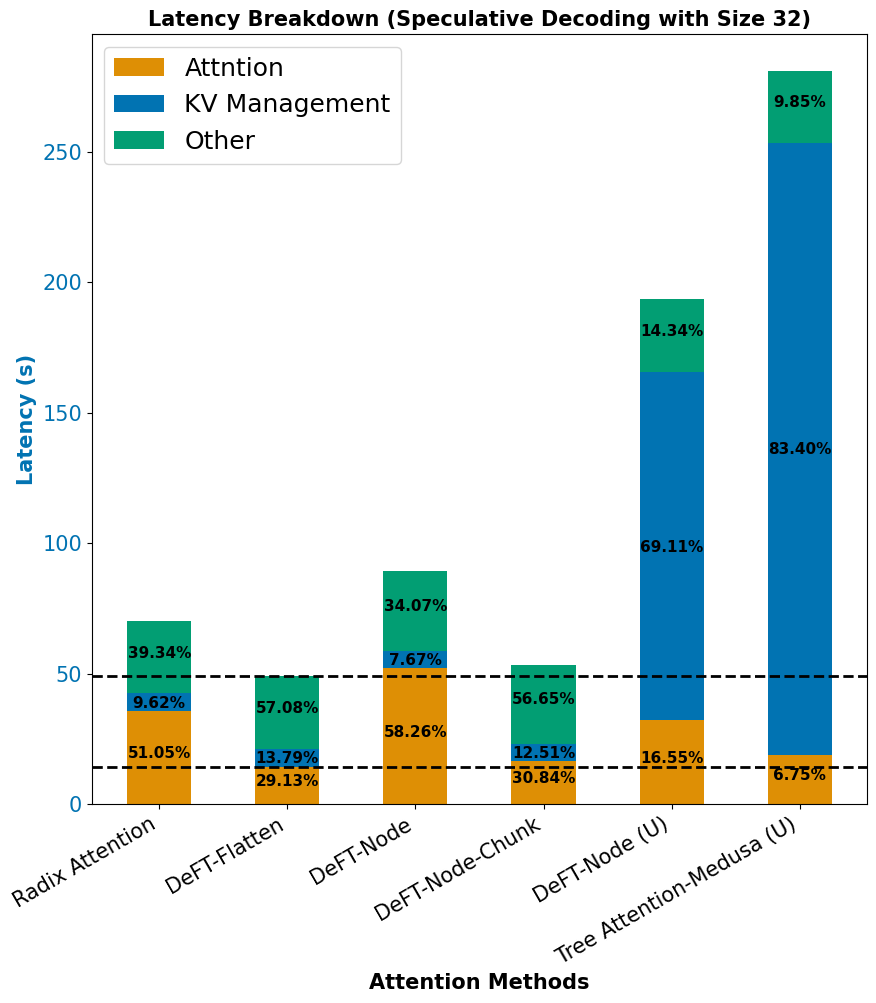}
    \vspace{-.9em}
    \caption{\small
        \textcolor{black}{\textbf{Latency breakdown for speculative decoding} with a token tree of 32 queries, whose tree topology is from Medusa~\citep{cai2024medusa}.}
        \textit{U} means unpaged memory.
    }
    \vspace{-3em}
    \label{LB_sd}
\end{wrapfigure}
\paragraph*{Bottlenecks and trade-offs.}
We provide support for \algopt and baselines with KV cache in memory management (unpaged or paged) according to their designs. We visualize the latency breakdown for (1) KV cache management, (2) attention, and (3) other operations (including MLP calculation) in Figure \ref{LB_sd}.
We observe that with unpaged KV cache management in tree-based decoding, the bottleneck (69.1-83.4$\%$) is the data movement required to materialize the KV cache.
However, when we use paged memory management, attention becomes the new bottleneck (51.1-58.3$\%$), especially when the token tree is large.

\subsection{Decoding Behaviors: Latency and IOs.}
We evaluate \algopt's performance on various tree-based decoding tasks by measuring decoding latency (\autoref{tab:e2e_comp}), which  demonstrates \algopt's acceleration of tree attention and wall clock time. See attention latency (\autoref{tab:e2eAttn_comp}), IO (\autoref{tab:PIO_comp}), and inference accuracy (\autoref{tab:infer_acc}) in Appendix \ref{A:AEXP}.
\looseness=-1
Notably, \textbf{decoding latency} essentially represents the optimal end-to-end (e2e) latency, excluding other overheads such as prefill latency (which accounts for approximately 5-10\% of e2e latency) and framework-induced overheads (roughly 10-15\% of e2e latency), including tree search, branching, etc. We exclude these overheads because they are consistent across all baselines and to eliminate the influence of the framework.
\looseness=-1

\textbf{For few-shot prompting tasks}, we used a prompt with 4k tokens and performed 400 decoding iterations, achieving a 1.33$\times$ decoding speedup thanks to 1.70$\times$ faster attention calculation and an approximately 90\% reduction in IO.

\textbf{For speculative decoding tasks}, \algopt-Flatten achieved up to a 2.23$\times$ decoding speedup due to up to a 3.59$\times$ attention speedup, as the entire token tree (all queries) can share IO of the long prefix.
\looseness=-1

\textbf{For multi-step reasoning tasks}. While \algopt-Flatten improves attention speed by up to $1.36\times$, the decoding acceleration is less pronounced due to two reasons:(1) The narrow tree width (10) restricts KV cache reuse, though increasing it to 50 in few-shot prompting significantly boosts decoding speed ($1.2\times$-$1.5\times$ over 100 iterations, see Appendix \ref{A:AEXP}). (2)The small number of tokens in the tree keeps attention at ~30\% of decoding latency, compared to 50-80\% in speculative decoding. A longer prompt length increases attention computation overhead, leading to greater speedup, as shown in \autoref{tab:ablation_promptlen2}.

\subsection{Ablation Study\label{subsec:ablation}}
We evaluate the influence of different KV splitting strategies, model sizes, and prompt lengths in this subsection. See more ablations in Appendix \ref{A:exp}, including the influence of different GPUs (\autoref{tab:e2eAttn_4090}), chunk sizes during KV splitting (\autoref{fig:ablation_chunk}), and model architectures (\autoref{tab:e2eAttn_34b} and \autoref{tab:e2eAttn_7b}).

\paragraph{The impact of KV splitting strategy in \algopt.}
We compared three KV splitting strategies with the baseline Radix Attention, as shown in \autoref{tab:e2e_chunk}. \algopt-Flatten consistently outperforms the others across all tree-structured settings. \algopt-Node-Chunk generally performs better than \algopt-Node because it splits large nodes into smaller chunks for more balanced computations, especially when $b\leq30$ and $t\leq64$, as well as in reasoning tasks like Keyword and Set. However, it struggles with many small nodes (e.g., prompts with around 1k tokens), leading to slower inference due to more GPU execution rounds required for additional QKV groups (see $t=256$ for \algopt-Node-Chunk).

\begin{table}[!t]
    \setlength\tabcolsep{2pt}
    \centering
    \small
    \caption{\small
        \textbf{ [Different KV Splitting Strategies] Comparison of \algopt-Node, \algopt-Node-Chunk and \algopt-Flatten in average attention latency (second) with NVIDIA A100 (80GB) for Llama3-8B model(GQA).}
        This table is supplementary to \autoref{tab:e2eAttn_comp}. The fastest method is in \textbf{bold}, and the second fastest is \underline{underlined}. \textcolor{blue}{Radix Attention} is the best baseline in decoding latency. See details of more baselines in \autoref{tab:e2eAttn_comp}.
        \label{tab:e2e_chunk}
    }
    \vspace{-1.em}
    \begin{footnotesize}
        \resizebox{0.95\columnwidth}{!}{
            \begin{tabular}{ccccccccccccc}
                \toprule
                \multirow{2}{*}{Memory} & \multirow{2}{*}{Method}                  & \multicolumn{3}{c}{Few-shot Prompting }
                                        & \multicolumn{4}{c}{Multi-Step Reasoning}
                                        & \multicolumn{4}{c}{Speculative Decoding}
                \\ \cmidrule(lr){3-5} \cmidrule(lr){6-9} \cmidrule(lr){10-13}
                                        &                                          & \texttt{b=20}                           & \texttt{b=30}     & \texttt{b=50}
                                        & \textit{Sorting}                         & \textit{Document}                       & \textit{Keyword}  & \textit{Set}
                                        & \texttt{t=32}
                                        & \texttt{t=64}                            & \texttt{t=128}                          & \texttt{t=256}
                \\ \midrule

                \multirow{3}{*}{Paged}
                                        & \textcolor{blue}{Radix Attention}        & \underline{5.99}                        & \underline{7.30}  & \underline{9.96}
                                        & \underline{39.37}                        & \underline{24.69}                       & \underline{3.11}  & \underline{5.13}
                                        & 25.73                                  & 40.47                             & 76.10            & 145.43
                \\
                                        & \algopt-Node                             & 10.59                                   & 10.62             & 10.85
                                        & 42.96                                    & 33.29                                   & 6.16              & 9.58
                                        & 34.59                                  & 34.41                                & 34.96 & \underline{41.78}

                \\
                                        & \algopt-Node-Chunk
                                        & 8.52                                     & 9.69                                    & 13.45
                                        & 49.63                                    & 36.37                                   & 4.77              & 7.40
                                        & \underline{14.54}                        & \underline{20.28}                       & \underline{32.57}             & 57.26
                \\
                                        & \algopt-Flatten                          & \textbf{3.47}                           & \textbf{4.07}     & \textbf{5.87}
                                        & \textbf{28.41}                           & \textbf{21.45}                          & \textbf{2.57}     & \textbf{3.83}
                                        & \textbf{13.15}                           & \textbf{16.79}                          & \textbf{24.46}    & \textbf{40.56}

                \\

                \bottomrule
            \end{tabular}
        }

    \end{footnotesize}
\end{table}

\paragraph{The influence of prompt length.}
See \autoref{tab:ablation_promptlen2}.
With a longer prompt, \algopt-Flatten shows a more pronounced speedup in the same model, since the attention overhead is proportional to the token count in the decoding tree, while the FFN overhead remains nearly constant for the same model. See \autoref{abla:tpot}, \autoref{abla:decoding_latency} and \autoref{abla:attn_latency} for more results.

\paragraph{The influence of model size.}
See \autoref{tab:model_size1}.
With the larger model, Codellama-34B, \algopt-Flatten achieves slightly reduced but significant (up to 1.78$\times$) decoding speedup.
The performance reduction is attributed to a lower A/F-LR, as the FFN overhead is greater due to larger hidden dimensions.
\looseness=-1

\begin{table}[!t]
    \vspace{-1.0em}
    \setlength\tabcolsep{2pt}
    \begin{minipage}{0.35\textwidth}
        \centering
        \small
        \caption{\small
            \textbf{[Different Prompt Lengths] Comparison of \algopt-Flatten and Radix Attention in the efficiency of multi-step reasoning task \texttt{sorting}.}
            The original prompt length is approximately 1K tokens, and we pad it to lengths of 5K, 8K, or 10K tokens.
            \label{tab:ablation_promptlen2}
        }
        \vspace{-1.em}
        \resizebox{\textwidth}{!}{
            \begin{tabular}{ccccc}
                \toprule
                \multirow{2}{*}{Speedup} & \multicolumn{4}{c}{Prompt length L}
                \\ \cmidrule(lr){2-5}
                                         & \texttt{L=1k}                       & \texttt{L=5k} & \texttt{L=8k}
                                         & \texttt{L=10k}
                \\ \midrule
                \textbf{Attention}
                                         & 1.39$\times$                        & 1.71$\times$  & 1.97$\times$
                                         & 1.84$\times$
                \\
                \midrule
                \textbf{Decoding}
                                         & 1.09$\times$                        & 1.37$\times$  & 1.53$\times$
                                         & 1.67$\times$
                \\
                \bottomrule
            \end{tabular}
        }

    \end{minipage}%
    \quad
    \begin{minipage}{0.6\textwidth}
        \centering
        \caption{\small
            \textbf{[Different Model Sizes] Comparison of decoding latency speedup and Attention/FFN latency ratio (in short as \textcolor{red}{A/F-LR}) between \algopt and \textcolor{blue}{Radix Attention} for Codellama-34B and Codellama-7B models.}
            \textcolor{blue}{Radix Attention} is the best baseline in decoding latency.
            $b$ represents the tree width, and $t$ denotes the token tree size.
            For multi-step reasoning, we test the task \textit{sorting} whose prompt length is about 1k tokens.
            \looseness=-1
            \label{tab:model_size1}
        }
        \vspace{-1.em}
        \resizebox{\textwidth}{!}{
            \begin{tabular}{cc|c|c|c}
                \toprule
                \multirow{2}{*}{Metric} & \multirow{1}{*}{Model Size}
                                        & \multirow{2}{*}{\makecell{Few-shot Prompting                                                \\\texttt{b=30}}}
                                        & \multirow{2}{*}{\makecell{Multi-step Reasoning                                              \\ \textit{Sorting}}}
                                        & \multirow{2}{*}{\makecell{Speculative Decoding                                              \\ \texttt{t=64}}} \\

                \\
                \midrule
                \multirow{2}{*}{\makecell{Decoding Time                                                                             \\ Speedup}}
                                        & 7B                                             & 1.34$\times$ & 1.09$\times$ & 1.85$\times$ \\
                                        & 34B                                            & 1.23$\times$ & 1.03$\times$ & 1.78$\times$

                \\ \midrule
                \multirow{2}{*}{\makecell{Radix Attention's                                                                           \\ \textcolor{red}{A/F-LR}}}
                                        & 7B                                             & 1.27         & 1.12         & 2.12         \\
                                        & 34B                                            & 0.80         & 0.48         & 1.66

                \\ \midrule
                \multirow{2}{*}{\makecell{\algopt-Flatten's                                                                           \\ \textcolor{red}{A/F-LR}}}
                                        & 7B                                             & 0.68         & 0.89         & 0.69         \\
                                        & 34B                                            & 0.45         & 0.42         & 0.49

                \\
                \bottomrule
            \end{tabular}
        }
        \label{tab:infer_acc2}
    \end{minipage}%
    \vspace{-1.2em}
\end{table}

\section{Conclusion\label{Disc}}

We propose \algopt-Flatten, a hardware-efficient attention algorithm optimized for tree-structured LLM inference. It effectively addresses memory access and GPU utilization bottlenecks by reusing shared prefixes' KV cache and evenly distributing workload across partitions. \algopt-Flatten's key strengths lie in its prefix-sharing awareness and load balancing, making it versatile for various tree-structured tasks. It also scales well with larger search spaces and multiple branches. Our results show that \algopt-Flatten achieves up to $2.23\times$/$3.59\times$ speedup in decoding and attention latency, outperforming baselines in tasks such as few-shot prompting, multi-step reasoning, and speculative decoding. 
Our ablation studies highlight that: (1) balanced partitioning is critical, (2) \algopt-Flatten delivers significant speedups across various LLM models and GPU architectures, and (3) \algopt-Flatten provides even greater speedups with larger token sizes (e.g., longer prompts) and more branches in tree-structured requests.

\clearpage

\section*{Acknowledgement}
This work was supported in part by the National Science and Technology Major Project (No.\ 2022ZD0115101), Research Center for Industries of the Future (RCIF) at Westlake University, Westlake Education Foundation, and Westlake University Center for High-performance Computing.
\bibliography{sources/paper}
\bibliographystyle{configuration/iclr2025_conference}

\clearpage
\onecolumn
{
    \hypersetup{linkcolor=black}
    \parskip=0em
    \renewcommand{\contentsname}{Contents of Appendix}
    \tableofcontents
    \addtocontents{toc}{\protect\setcounter{tocdepth}{3}}
}

\appendix
\section{Appendix}
\subsection{Components of System Support for \algopt \label{comp_func}}

\begin{figure}[!h]
    \begin{center}
        \includegraphics[width=1.\textwidth,height=0.35\textheight]{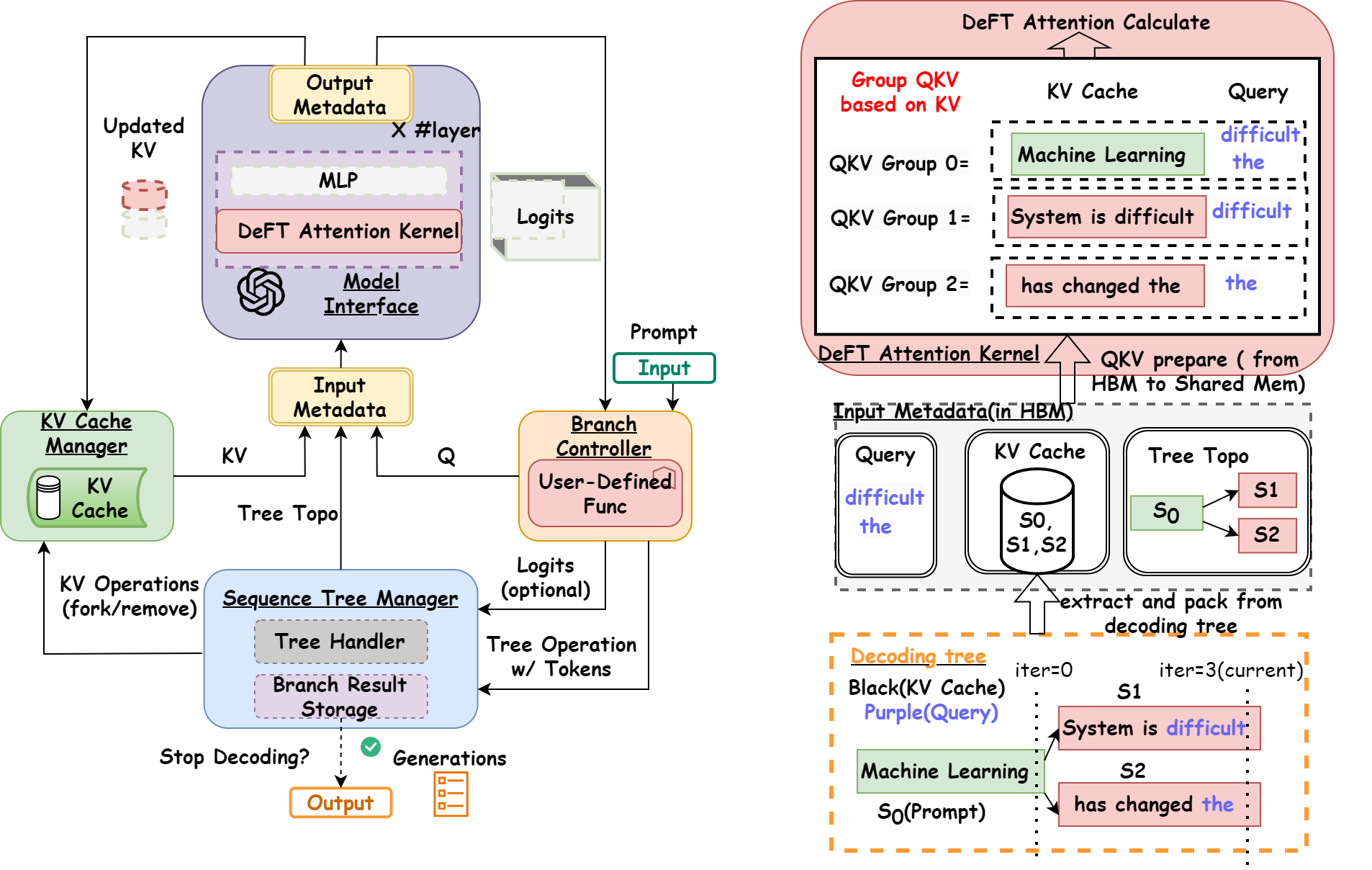}
    \end{center}
    \vspace{-1.em}
    \caption{\small
        \textbf{Illustration of \algopt}.
        (Left) System overview.
        (Right) The data flow of \algopt-Node (\algopt-Flatten is similar except for QKV partitioning) using a decoding tree example.
    }
    \label{fig:DeFT_sys}
    \vspace{-1.em}
\end{figure}

The left part of \autoref{fig:DeFT_sys} shows the coordination of different components for efficient and flexible tree-based decoding. The details of functions for system components of \algopt are as below:
\looseness=-1
\begin{enumerate}[nosep, leftmargin=12pt]
    \item \textbf{Branch Controller}:
          It makes the tree decoding process forced by a user-defined function (e.g. branch to two children every $3$ iterations, as the example shown in the right of \autoref{fig:DeFT_sys}).
          Tree-search-based algorithms can be applied here using the decoding tree's topology information.
    \item \textbf{Sequence Tree Manager}:
          It maintains the topology of the decoding tree based on the tree operations and tokens from the Branch Controller.
          The tree operations like pruning and branching will be executed by \textit{Tree Handler} in this component. \textit{Branch Result Storage} will record token generation results of all branches in the decoding tree, and output when the decoding stops.
    \item \textbf{\kvcache Manager}:
          It will maintain \kvcache with a tree structure.
          A map between sequence IDs in the decoding tree and \kvcache index is kept, which will be updated based on KV operations\footnote{
              e.g. when a node is pruned in the decoding tree, its KV space will be evicted using a \textit{Remove} operation.
          } from the Sequence Tree Manager. We provide both paged~\citep{kwon2023efficient} and unpaged memory management in this part to fit different attention kernels.
    \item \textbf{Model Interface}: pass input metadata to \algopt Attention kernel and MLP module, then return logits and memory pointers of updated KV cache.
\end{enumerate}

The right part of \autoref{fig:DeFT_sys} further showcases the key data flow of the system through a decoding tree example. For simplicity, we present \algopt-Node here and \algopt-Flatten is similar except for QKV partitioning. Input metadata will be extracted by three components we mentioned above, then loaded from HBM to shared memory in a group manner after the \textsc{QKV Preparation Phase} discussed in Section \ref{sec:DeFT_Attn}.
Then QKV groups will be processed by \textsc{\algopt Attention Kernel} in \textsc{Attention Calculation Phase} of \algopt. See details of techniques in these two phases in Appendix \ref{A:Tech}.

\subsection{Discussion of Tree-based Decoding\label{Tree_decoding_disc}}
\begin{figure}[!ht]
    \centering
    \subfloat[\small
        (Left) Sequence KV with queries in a tree for parallel decoding \citep{miao2023specinfer,cai2024medusa}, where a \textit{causal mask}  is applied to record the causal information among queries in a tree of tokens. (Right) Tree KV with parallel queries for shared prefixes in multi-step reasoning.
        \looseness=-1
        \label{td_multiturns}
    ]{\includegraphics[width=0.7\linewidth]{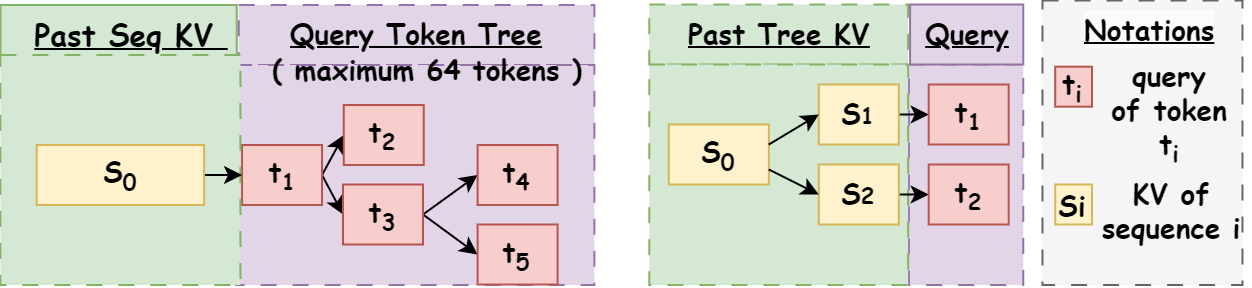}}
    \hfill
    \subfloat[\small
        \textit{Bit Mask} in SpecInfer~\citep{miao2023specinfer} to record the causal information between query tokens in a tree structure. The decoding tree is in the left part of \autoref{td_multiturns}.
        \looseness=-1
        \label{td_parallel}
    ]{\includegraphics[width=0.6\linewidth]{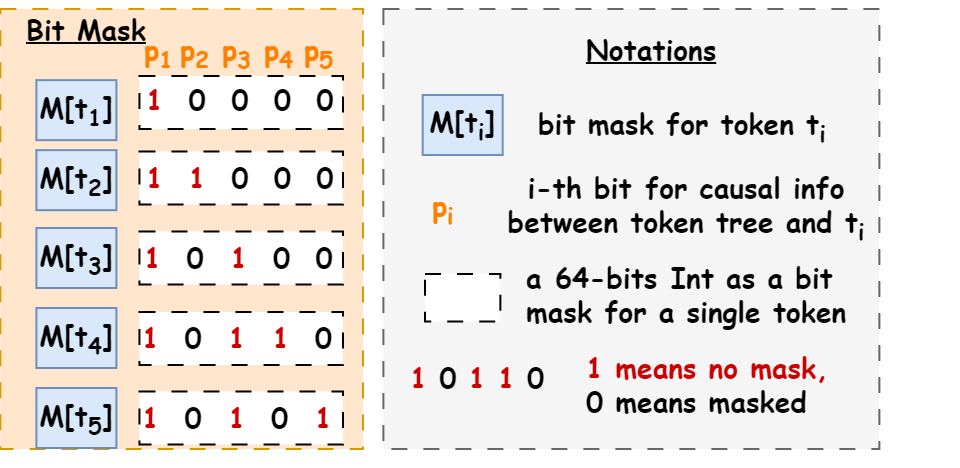}}
    \vspace{-0.8em}
    \caption{ \small
        Discussion of \textbf{tree-based decoding} with tree queries \citep{miao2023specinfer} and tree KV.\label{TD_disc}
        \looseness=-1
    }
    \vspace{-1em}

\end{figure}
Tree-based decoding could have tree-structured \kvcache for storage with awareness of shared prefixes \citep{zheng2023efficiently}, or tree-structured queries in parallel/speculative decoding \citep{miao2023specinfer,cai2024medusa}, as shown in \autoref{TD_disc}.  A general decoding could both do with tree KV and tree queries, which could reduce redundancy (e.g. IO, storage, computation, etc) of shared prefixes, as well as increase the generated tokens per decoding iteration.

The existing inference frameworks~\citep{zheng2023efficiently, gim2023prompt} focused on tree-based decoding efficiency primarily aim to: (1) reduce memory footprints~\citep{zheng2023efficiently} to enable larger batch sizes for higher throughput; (2) reuse the prompt cache~\citep{gim2023prompt} to avoid recomputation of the KV cache for faster time-to-first-token (TTFT). However, their designs do not specifically target reducing the latency of the entire decoding process. We observe that the tree-structured feature of LLM inference could provide us with some advantages to speed up the decoding itself.

\paragraph*{Analysis of speedup potential in tree-based decoding.
} In tree-based decoding, KV cache and queries can be structured in a tree. Not only can we store KV cache in a tree, but also we can load QKV with awareness of tree topology during attention calculation, to minimize the expensive IO between HBM and on-chip shared memory of GPUs. We explain it in two case studies of complex scenarios with tree-structured interactions: (1) multi-step reasoning~\citep{yao2023tree,xie2024self}; (2) speculative decoding~\citep{cai2024medusa,miao2023specinfer}.

\begin{figure}[!ht]
    \begin{center}
        \includegraphics[width=0.7\columnwidth]{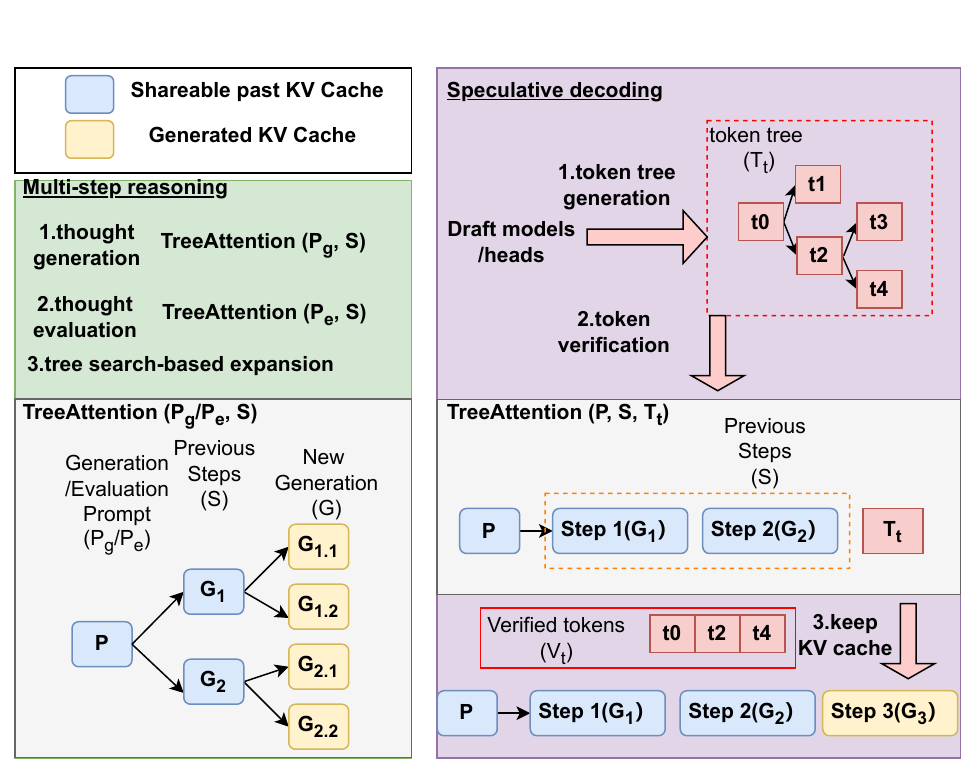}
    \end{center}
    \vspace{-1em}
    \caption{\small Analysis for two case studies of tree-based decoding. (Left) Multi-step reasoning. (Right) Speculative decoding. Blue boxes mean shareable past KV cache in storage and memory access during the tree attention calculation, while yellow boxes mean the KV cache of generated context. \label{fig:treeAna} }

\end{figure}

\paragraph*{Case study 1: multi-step reasoning.} As shown in the left part of \autoref{fig:treeAna}, we can summarize process of multi-step reasoning~\citep{hao2023reasoning,yao2023tree,besta2023graph} to three phases:  (1) \textit{Thought Generation}: generate k candidates for the next thought step based on a generation prompt $P_g$ and previous steps $S$; (2) \textit{Thought Evaluation}: When presented with a frontier of various thoughts, a LLM as state evaluator measures previous thoughts $S$ based on an evaluation prompt $P_e$ towards resolving the problem. This assessment acts as a heuristic for the search algorithm, guiding it on which states to pursue further and the sequence in which to explore them; (3) \textit{Tree Search-based Expansion}: play different search algorithms~\citep{lu2022neurologic,liu2023making,xie2024self} to explore search space, which influences the future tree topology. In both (1) and (2), we can share IO of KV cache for $P_g$/$P_e$ and $S$ during tree attention calculation.

\paragraph*{Case study 2: speculative decoding.} As shown in the right part of \autoref{fig:treeAna}, we can summarize the process of speculative decoding~\citep{cai2024medusa,miao2023specinfer} to tree phases: (1) \textit{Token Tree Generation}: multiple small draft models~\citep{miao2023specinfer} or fine-tuned heads~\citep{cai2024medusa} generate multiple sequences of tokens based on prompt $P$, then they are merged to a speculated token tree $T_t$, which is very fast (e.g. $1\%$ of time overhead in SpecInfer~\citep{miao2023specinfer}); (2) \textit{Token Verification}: based on these tree-structured token candidates $T_t$, verify the correctness of its tokens against an LLM's output, where tree-attention calculation is the bottleneck of the process~\citep{miao2023specinfer}. In (2), we can share IO of KV cache for $P$ and $S$ during tree attention calculation.

\paragraph*{Why existing tree-attention algorithms are not enough?} The existing tree-attention algorithms are either in-efficient in memory access~\citep{cai2024medusa,miao2023specinfer} or not suitable for general tree-based decoding~\citep{miao2023specinfer} with more than 64 tokens in the token tree.
\begin{itemize}[nosep, leftmargin=12pt]
    \item In SpecInfer\citep{miao2023specinfer}, as shown in Figure \autoref{td_parallel}, a \textit{bit mask} is utilized to record the causal information among queries of a token tree. Each token $t_i$ in queries will have a 64-bit Int as a \textit{bit mask}, where j-th bit means the causal relationship between query of $t_i$ and \kvcache of $t_j$. The advantage of this mask design is that it greatly reduces IO, but it results in the maximum number of tree tokens being only 64, which is not practical for scenarios with tree-structured \kvcache. What's more, it is not IO-aware for KV cache as it will load KV cache of the entire tree for each query.
    \item Medusa~\citep{cai2024medusa} is suitable for general tree-based decoding, but it is not hardware-efficient due to significant IOs of a dense causal mask and partial results during attention calculation (e.g. softmax).
\end{itemize}

\subsection{Discussion of Concurrent Works\label{Concurrent_disc}}
There are some concurrent works~\citep{athiwaratkun2024bifurcated,ye2024chunkattention,juravsky2024hydragen,zhu2024relayattention} in attention algorithm design for single-context large-batch sampling, where the goal is to generate multiple sequences from a single context(e.g. system prompt or few-shot examples), which is a special case  of tree-based decoding with a depth of 1. The design of their algorithms are based on this feature, which means they can not suit well in attention calculation of a tree with more than two levels of prefixes with efficiency.
\paragraph*{Insights and techniques in common.} Both concurrent works and \algopt have the insight that memory access is the bottleneck of LLM inference, and decomposing attention across subsequences to reduce the memory access of the prefix KV: (1) calculate attention $A_p$, $A_s$ over prefix and suffixes, respectively; (2) get finial attention by online softmax merging~\citep{dao2022flashattention,dao2023flashdecoding} based on $A_p$ and $A_s$. Here are the details of the correctness proof:

\begin{itemize}[nosep, leftmargin=12pt]
    \item Let's say we have key tensor $K \in R^{(l_{kv},d)}$,  value tensor $V \in R^{(l_{kv},d)}$, and query tensor $Q \in R^{(l_q,d)}$. Consider the general case K and  V are partitioned across the sequence (row) dimension into two parts for prefix and suffixes, respectively: $K=K_p \parallel K_s$, and $V=V_p \parallel V_s$, with $\parallel$ denoting concatenation along the row axis.
    \item We calculate the attention $A_p$, $A_s$ over prefix and suffixes, where
          $
              A_p = \langle Q, K_p, V_p \rangle, \quad A_s = \langle Q, K_s, V_s \rangle,
          $
          and
          $
              \langle q, k, v \rangle = \operatorname{Softmax}\left(\frac{q k^T}{\sqrt{d}}\right) v.
          $
          \item Based on \autoref{eq:seg_softmax}, we can have segmented attention $\langle \mQ, \mK, \mV \rangle =\operatorname{SegAttn}(\mA_p, \mA_s)$.
\end{itemize}

\begin{table}[!ht]
    \setlength\tabcolsep{2pt}
    \centering
    \small
    \caption{\small Comparison among \algopt and concurrent works in single-context large-batch sampling scenarios, including Chunk-Attention~\citep{ye2024chunkattention}, Hygragen~\citep{juravsky2024hydragen} and Bifurcated-Attention~\citep{athiwaratkun2024bifurcated}. RelayAttention~\citep{zhu2024relayattention} and Cascade-inference~\citep{cascadeinference} are similar to Hygragen. More $\star$ means more balanced workloads after tree split, which also shows how insensitive the acceleration is to the tree topology. \label{tab:concurrent}}
    \begin{footnotesize}
        \resizebox{1.0\columnwidth}{!}{
            \begin{tabular}{@{}ccccccc@{}}
                \toprule
                \textbf{Method}                    & Chunk-Attention
                                                   & Hygragen
                                                   & Bifurcated-Attention & \algopt-Node  & \algopt-Node-Chunk
                                                   & \algopt-Flatten                                                  \\
                \midrule
                \textbf{IO-aware levels}           & 2 (depth $\leq$ 1)  & 2 (depth $\leq$ 1)  & 2 (depth $\leq$ 1)    & all (every depth) & all (every depth) & all (every depth)           \\
                \textbf{Tree KV split granularity} & by node first, then by block & by tree depth & by tree depth   & by tree node  & by tree node, then by block & flatten tree, then by block \\
                \textbf{Load-balanced level}       & $ \star \star \star   $   & $\star\star $       & $\star\star $  & $\star$  & $\star\star\star $  & $ \star \star \star \star $       \\
                \textbf{Goal metrics}              & throughput         & throughput    & latency  & latency  & latency   & latency                 \\
                \bottomrule
            \end{tabular}
        }
    \end{footnotesize}
    \vspace{-1.25em}
\end{table}

\paragraph*{Comparison of differences.} The existing works of single-context large-batch sampling are not hardware-efficient for general tree-based decoding for two reasons, as shown in \autoref{tab:concurrent}:
\begin{itemize}[nosep, leftmargin=12pt]
    \item They are designed for decoding trees with only two levels—prefixes at the root and suffixes at depth 1. For decoding trees with multiple levels of prefixes, their algorithm can only reduce the IO of the prompt at the root of the tree. However, in scenarios such as multi-step reasoning~\citep{yao2023tree,besta2023graph,hao2023reasoning}, the token length of non-root prefixes can also be very long (e.g., thousands of tokens), and their KV cache's IO is not reused. \algopt can reuse KV IO of all non-leaf prefixes in a general decoding tree, providing greater acceleration potential.
    \item They have not addressed the unbalanced workload problem in tree-based decoding. Nodes in the decoding tree can vary significantly, making it crucial to split the tree and group QKV in a way that ensures balanced calculations for each QKV group. Simply dividing based on depth alone is insufficient. For example, in speculative decoding, the prefix might contain thousands of tokens, while each layer only processes a few dozen tokens~\citep{cai2024medusa,miao2023specinfer}. 
\end{itemize}

\subsection{Discussion of Techniques in Efficient Attention Algorithm Design\label{A:Tech}}

\begin{figure}[!ht]
    \begin{center}
        \includegraphics[width=1.\columnwidth]{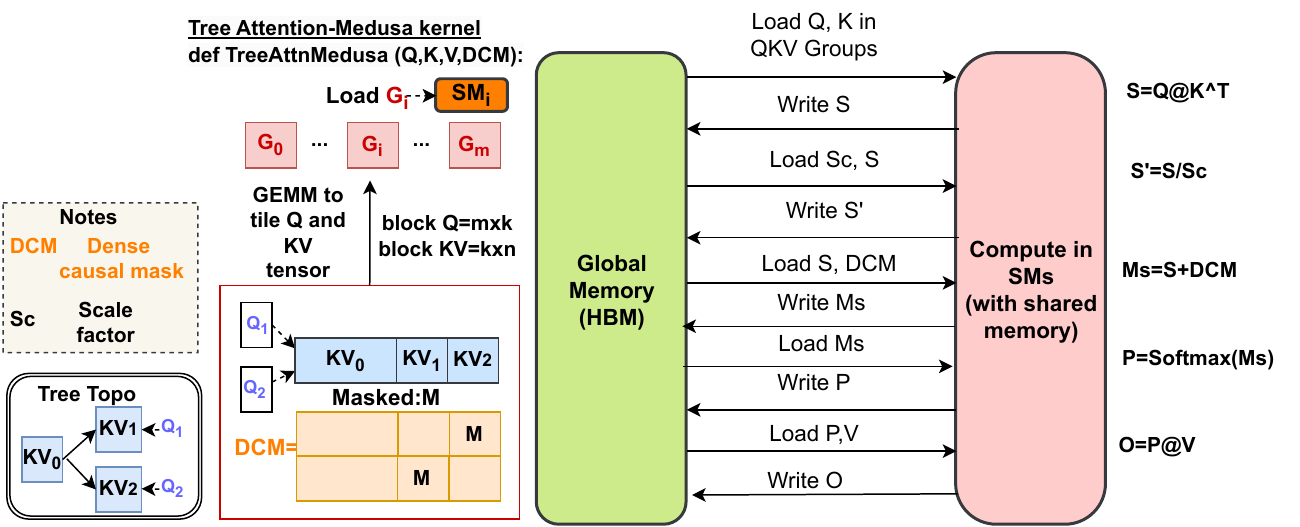}
    \end{center}
    \vspace{-1em}
    \caption{\small Operations of \treeAttn\citep{cai2024medusa}. No \textit{Kernel Fusion} or \textit{Tiling} strategy is applied, which introduces significant IO of partial results like $\mQ \mK^\top$, DCM, and $\text{Softmax}$ between GPU global memory and on-chip shared memory.
        \label{fig:CMA_AttnOP} }
\end{figure}

\begin{table}[!ht]
    \setlength\tabcolsep{2pt}
    \centering
    \small
    \caption{\small Technique list of \algopt. What we propose is in \textcolor{red}{red}. The details of the first four techniques are in Section \ref{sec:DeFT_Attn}, while the details of the following techniques are discussed in this chapter.
        \label{tab:techlist}}
    \begin{footnotesize}
        \resizebox{1.0\columnwidth}{!}{
            \begin{tabular}{@{}cc@{}}
                \toprule
                \textbf{Technique}                                                         & \textbf{Goal}                                                                        \\
                \midrule
                \textcolor{red}{\textit{KV-Guided Grouping}}               & High utilization of GPU and minimal KV cache IO between HBM and shared memory.
                \\
                \textcolor{red}{\textit{Flattened Tree KV Splitting}}                       & Balanced attention calculation for high computation efficiency.
                \\
                \textit{Bit Causal Mask}~\citep{miao2023specinfer}                          & Record causal information of tokens in the decoding tree with little IO cost.
                \\
                \midrule
                \textit{Kernel Fusion}  ~\citep{dao2022flashattention,dao2023flashdecoding} & Reduce partial results IO (e.g. $\mQ \mK^T$,  Mask $M$, and $\text{Softmax}$, etc ).
                \\
                \textit{Tiling} ~\citep{dao2022flashattention,dao2023flashdecoding}         & Enable attention calculation within limited size of GPU's shared memory.
                \\
                \textcolor{red}{\textit{Tree-topology Aware Global Reduction}}             & To get the correct tree attention of the entire decoding tree.

                \\
                \bottomrule
            \end{tabular}
        }
    \end{footnotesize}
    \vspace{-1.25em}
\end{table}

In this subsection, we summarize and discuss the common techniques in existing designs of efficient attention algorithms and kernels : (1) \textit{Kernel Fusion} with \textit{Tiling} strategy \citep{dao2022flashattention,hong2023flashdecoding,miao2023specinfer}; (2) \textit{Tree-topology Aware Causal Mask} \citep{miao2023specinfer,cai2024medusa}; (3) \textit{KV Split} with \textit{Global Reduction}\citep{hong2023flashdecoding}. Then we explain the details of design in \algopt Attention Kernel, where the techniques are in \autoref{tab:techlist}.

\textit{Kernel Fusion} is a common technique of IO reduction: if multiple operations are performed on the same input, it is more efficient to load the input once from HBM rather than loading it multiple times for each operation; Similarly, the same principle applies when transferring output from shared memory to HBM.
To fuse all the attention operations into one GPU kernel with the limited size of shared memory, we further utilize the commonly employed \textit{Tiling} strategy~\citep{dao2022flashattention,dao2023flashdecoding}: split queries and \kvcache within each QKV group to small blocks to prevent materialization of attention matrix in HBM by computing attention within the limited size of shared memory, then incrementally performing the softmax reduction as the formulation in \autoref{eq:seg_softmax} to reconstruct the attention.

\begin{remark}[Importance of tiling and fused kernel during \textsc{Attention Calculation Phase}]
    Methods in this phase can be roughly divided into two categories: (1) without tiling and kernel fusion: Tree Attention in Medusa \citep{cai2024medusa}, which introduces significant IO operations for partial results (i.e.. $\mQ \mK^\top$ and $\text{Softmax}$), as shown in \autoref{fig:CMA_AttnOP}; (2) with tiling and a fused kernel: Flash Decoding \citep{dao2023flashdecoding}, Tree Attention in SpecInfer \citep{miao2023specinfer} and our \algopt.

\end{remark}

\begin{figure}[!ht]
    \begin{center}
        \includegraphics[width=1.\textwidth,height=0.15\textheight]{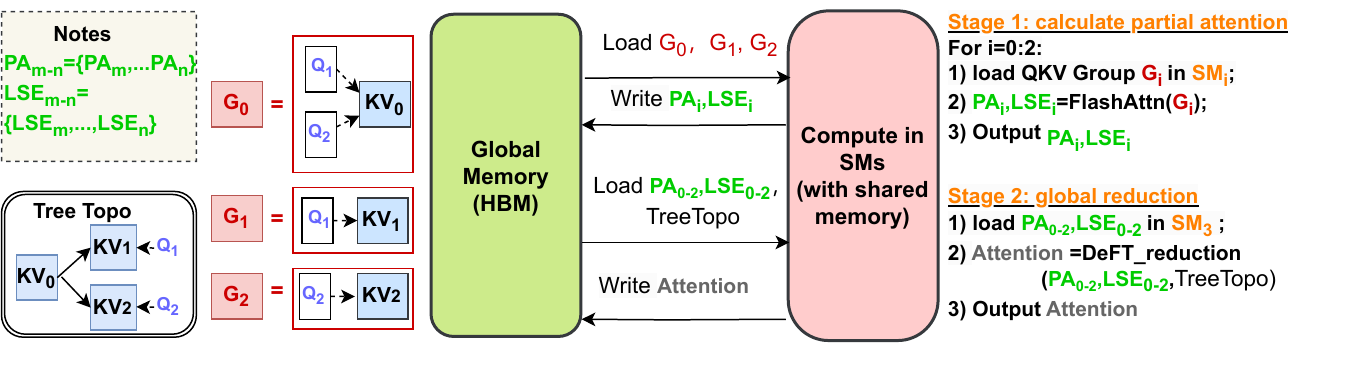}
    \end{center}
    \vspace{-1.5em}
    \caption{\small \textit{Overview of two stages in \algopt Attention Kernel (\algopt-Node for example, and \algopt-Flatten is similar).} \textbf{Stage 1--calculate partial attentions.}
        Based on the QKV grouping results after \textsl{KV-Guided Grouping Strategy with Tree Split} as mentioned above, each QKV group ($G_{i}$) will be allocated to a thread block for Flash Attention~\citep{dao2022flashattention} calculation with common \textit{Kernel Fusion} and \textit{Tiling} strategy. Similar to \flashdecoding~\citep{dao2023flashdecoding}, we not only get partial attention ($PA_{i}$) but also return ``LogSumExp'' ($LSE_{i}$) as a weight parameter for the next stage's reduction.
        \textbf{Stage 2--global reduction.}
        Upon receiving $PA_{i}$ and $LSE_{i}$ for each QKV group $G_i$, \algopt now performs a \textsl{Tree-Topology-Aware Global Reduction} ($DeFT\_reduction$).
        Guided by the tree topology among sequence nodes of KV in the decoding tree, \algopt logically remaps the partial results of attention and LogSumExp to get the correct final attention for each query after reduction.
        The decoding tree is the same as the one in the left of \autoref{fig:QKV_prepare}. $SM_{i}$ means the streaming multiprocessor $i$ in GPU.
        \looseness=-1
        \label{fig:DeFT_2PKernel}
    }
\end{figure}

The \textit{Tree-topology Aware Causal Mask} (\textit{Causal Mask} for short) is introduced in speculative decoding works \citep{miao2023specinfer,cai2024medusa} to facilitate the calculation of attention for all queries within a decoding tree using a single GPU kernel. It achieves this by recording the causal relationships among queries and \kvcache in the decoding tree. As depicted in \autoref{TD_disc}, while originally designed for tree-based decoding with \kvcache for a sequence of tokens and tree-structured queries, the \textit{Causal Mask} can also be adapted to tree decoding with tree-structured \kvcache and parallel queries—a configuration targeted by \algopt to enhance efficiency.

\begin{remark}[The effects of introducing a causal mask]
    Causal mask brings two parts of redundancy:
    \begin{itemize}[nosep, leftmargin=12pt]
        \item Memory Access.
              Medusa~\citep{cai2024medusa} materializes the dense causal mask (DCM) in HBM to record the causal information between $n_q$ tokens in queries and $n_{kv}$ tokens in the \kvcache, thereby introducing a significant IO cost for loading this $n_q \times n_{kv}$-sized mask to shared memory.
              SpecInfer~\citep{miao2023specinfer} introduces a 64-bit integer as a bit causal mask (BCM) to record the causal information among up to 64 tokens, which incurs minimal IO cost from HBM to shared memory but is not suitable for decoding trees with more than 64 tokens.
              Details regarding the design of the bit mask in SpecInfer are discussed in Appendix \ref{Tree_decoding_disc}.
        \item Computation.
              In addition to the computational cost of generating the causal mask itself, there is an additional redundancy in computation: many of the matrix multiplication results of $\mQ \mK^\top$ are masked out and never utilized. Both Medusa and SpecInfer have this issue.
    \end{itemize}

    \algopt-Flatten in Appendix \ref{A:DeFT-Sub-alg} adopts a bit causal mask insipred by SpecInfer~\citep{miao2023specinfer} to minimize the IO of the causal mask. Details of the bit mask design are on the left of \autoref{fig:QKV_prepare}.

\end{remark}

\textit{Splitting} is introduced to improve GPU utilization in sequence-based decoding~\citep{hong2023flashdecoding}, which is necessary when the parallelism is limited by a small batch size for long-context scenarios.  \flashdecoding splits long KV and group QKV based on Q first, then these groups will be allocated to different streaming multi-processors (SMs) in the GPU to get partial attention via Flash Attention~\citep{dao2022flashattention}.

\begin{figure}[!ht]
    \centering
    \subfloat[
        Left: Illustration of \algopt Attention Kernel with two stages. Right: Global reduction kernel called in \algopt stage 2 illustrated in Figure~\ref{subfig:Stage2DeFT}.
        QKV Groups $G_0$,$G_1$ and $G_2$ are from \algopt QKV groups in \autoref{fig:QKV_prepare}.\label{subfig:DeFT_kernel}
    ]{\includegraphics[width=1.\columnwidth,height=0.2\textheight]{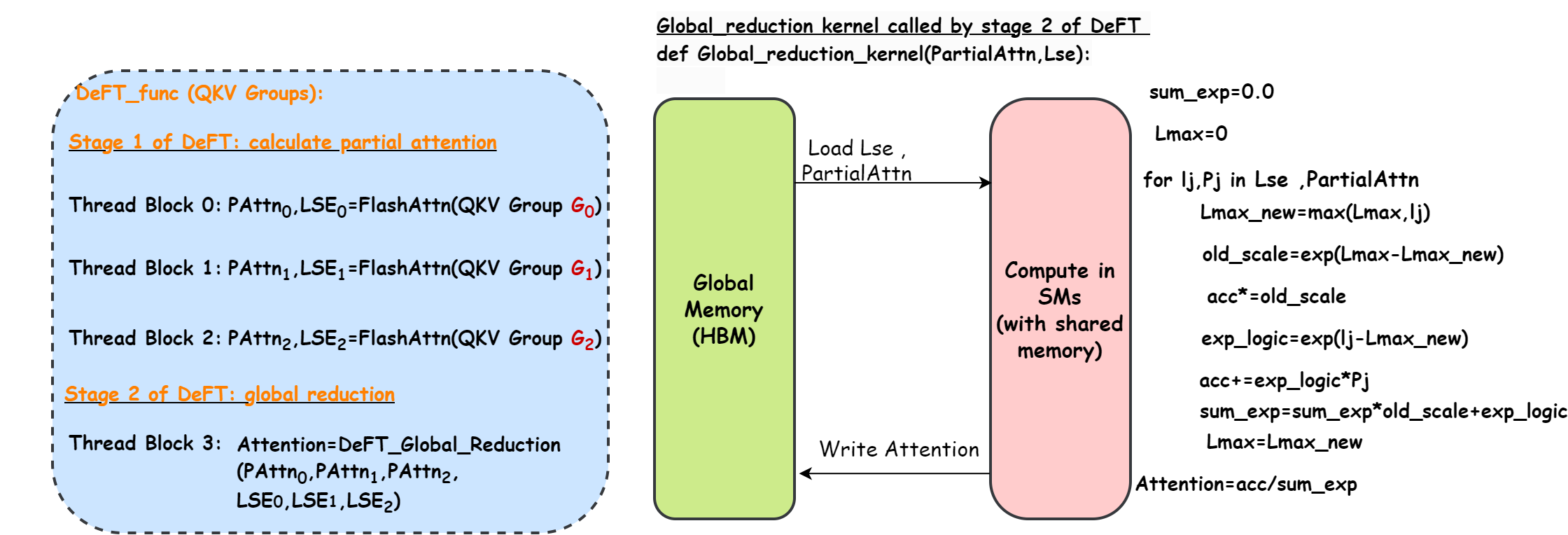}} \\
    \subfloat[
        Stage 2 of \algopt: Global Reduction (\algopt-Node for example).
        Based on tree topology in \autoref{fig:QKV_prepare}, we can group LogSumExp and Partial Attention based on Query, then we call the Global reduction kernel in the right of Figure~\ref{subfig:DeFT_kernel} to get the final attention.
        \looseness=-1
        \label{subfig:Stage2DeFT}
    ]{\includegraphics[width=1.\columnwidth,height=0.2\textheight]{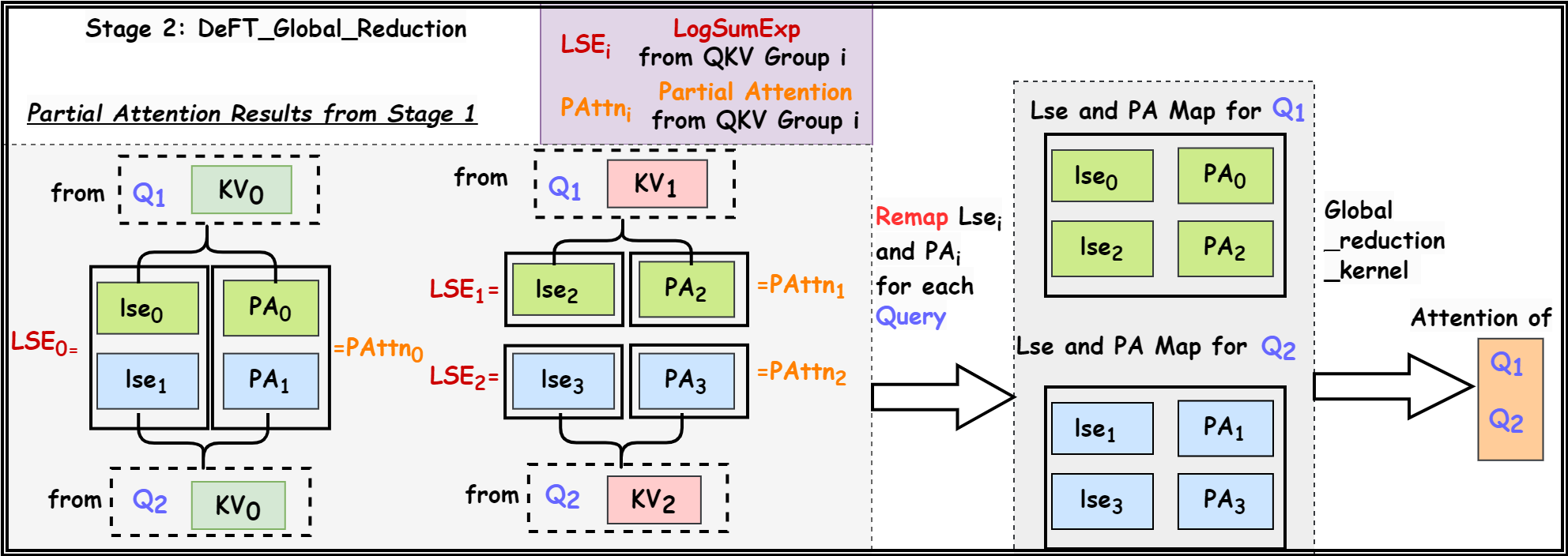}} \\
    \caption{\small \textbf{Detailed attention operations of \algopt kernel (\algopt-Node for example, and \algopt-Flatten is similar)}. Based on the same decoding tree in  \autoref{fig:QKV_prepare}. \label{fig:DeFT_kernel}}
\end{figure}
To obtain the accurate final attention, partial attentions from QKV groups with identical queries need to be grouped for \textit{Global Reduction}.

Similarly, \algopt also splits the decoding tree to different QKV groups for balanced workloads of SMs in the GPU, which is the \textsl{Flattened Tree KV Splitting} strategy we propose in Section \ref{sec:DeFT_Attn}, as illustrated in the bottom right part of \autoref{fig:QKV_prepare}. To obtain the correct tree attention, \algopt also requires a global reduction. However, the global reduction in \flashdecoding is for sequence-based decoding, which cannot aware the tree-topology for global reduction in tree-based decoding. Therefore, we propose \textsl{Tree-Topology-Aware Global Reduction}, as shown in the Figure \autoref{subfig:Stage2DeFT}.

Based on the techniques mentioned above, we designed the \algopt{} Attention Kernel with two stages, as shown in \autoref{fig:DeFT_2PKernel}, to execute the attention operations after the \textbf{QKV Preparation Phase} of \algopt, which we elaborated on in Section \ref{sec:DeFT_Attn}. For more details on the \algopt{} Attention Kernel, see \autoref{fig:DeFT_kernel}. The attention operations of \algopt{}-Flatten are omitted because they are very similar to those of \algopt{}-Node, except for the usage of the bit causal mask for tree attention calculation.

\subsection{Analysis: IO Complexity of \algopt\label{IOAna}}
This section analyzes the IO complexity of \algopt, showing a significant reduction in HBM accesses compared to existing attention algorithms.
Note that it is non-trivial to summarize the IO cost of the entire tree decoding process, thus we only compare IOs based on the decoding tree snapshot in a single iteration.

Consider a decoding tree with the features outlined in \autoref{tab:notations}, and we summarize the corresponding IO breakdown in \autoref{tab:io_breakdown}.

\begin{table}[!h]
    \centering
    \caption{\small \textbf{Notations}.}
    \label{tab:notations}
    \vspace{-1em}
    \bgroup
    \def\arraystretch{1.5}
    \begin{tabular}{p{0.5in}p{4.in}}
        \midrule
        $\displaystyle l_n$      & Number of leaf nodes in a decoding tree, which means how many queries are in this decoding iteration.                                                      \\
        $\displaystyle N_i$      & Total token length from the root node to leaf node i.                                                                                                      \\
        $\displaystyle N_{tree}$ & Total token length the entire tree.                                                                                                                        \\
        $\displaystyle \#node$   & Total number of nodes in entire tree.                                                          \\
        $\displaystyle n_i$      & The token length of node i.                                                              \\
        $\displaystyle d_{head}$ & Head dimension of LLM.                                                                                                                                     \\
        $\displaystyle s_{c}$    & Scale factor for scaled dot-product attention, typically denoted as $\sqrt{d_{\text{head}}}$.                                                              \\
        $\displaystyle F_{s}$    & Shared factor of reusing prefixes in tree attention, which means to which extent we can reduce IOs of KV cache: $F_s=(\sum_{i=1}^{ln} N_{i}) / N_{tree} $. \\
        \bottomrule
    \end{tabular}
    \egroup
\end{table}

It can be observed that \emph{due to the lack of tree-topology awareness, sequence-based decoding methods, such as naive attention and \flashdecoding, incur $F_s$ times more memory access overheads for \kvcache compared to \algopt-Node/Node-Chunk/Flatten and \treeAttn~\citep{cai2024medusa}}.

\begin{figure}[!ht]
    \begin{center}
        \includegraphics[width=1.\columnwidth]{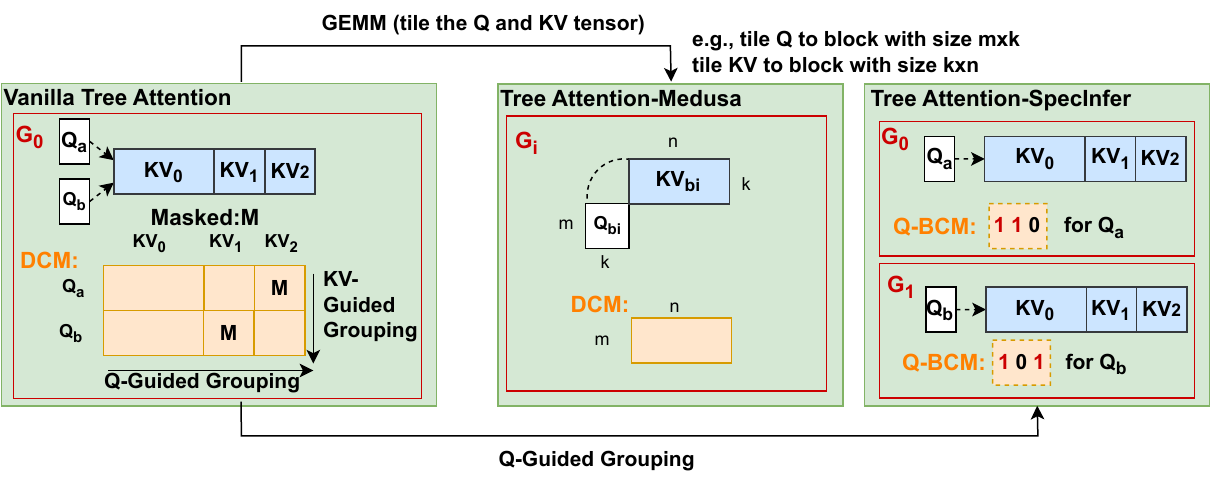}
    \end{center}
    \vspace{-1em}
    \caption{\small \textbf{(Supplementary to \autoref{fig:QKV_prepare}) QKV partitioning of Tree Attention~\citep{cai2024medusa,miao2023specinfer} and memory access.} Tree Attention-Medusa~\citep{cai2024medusa} partitions QKV by General Matrix Multiply(GEMM) in PyTorch. Tree Attention-SpecInfer~\citep{miao2023specinfer} adopts Q-Guided Grouping.  Q-$BCM$ is the Q-Guided bit causal mask for SpecInfer, where each bit represents the causal relationship between a query and a segment of tokens in the \kvcache. For example,Q-$BCM$ for $Q_a$ is "110" which means the first two segments of \kvcache $KV_0$ and $KV_1$ is valid for $Q_a$'s attention. The $Q_i$ and $K_j$ in the figure are the same as the ones in \autoref{fig:QKV_prepare}.
        \label{fig:treeattn_IO} }
\end{figure}

However, Tree Attention-Medusa entails higher IO overheads for partial results like $\mQ \mK^\top$ and $\text{Softmax}$ due to the lack of tiling and kernel fusion\footnote{
    Note that $\mQ \mK^T$, $\frac{\mQ \mK^\top}{s_{c}}$, $\mM+\frac{\mQ \mK^\top}{s_{c}}$ and $\text{Softmax}$ will load and write, so the IO cost contains a round-trip of memory access between HBM and shared memory, as shown in \autoref{fig:CMA_AttnOP}.
}.
What's more, a dense mask is introduced to record the causal information of tokens in the tree, with significant IO costs, as shown in the left of \autoref{fig:treeattn_IO}.

When the number of leaf nodes/queries $ln$ is sufficiently large, the IO cost of partial results might become comparable to that of the \kvcache.
For instance, in the Llama models~\citep{touvron2023llama,touvron2023llama2}, where $d_{head}\!=\!128$, with $l_n\!=\!29$, the total IO cost of $\mQ \mK^T$, $\mM$, $\frac{\mQ \mK^\top}{s_{c}}$, $\mM+\frac{\mQ \mK^\top}{s_{c}}$, and $\text{Softmax}$ matches that of the \kvcache.

\begin{table}[!t]
    \footnotesize
    \renewcommand{\arraystretch}{1.5}
    \centering
    \setlength{\tabcolsep}{4pt}
    \caption{\small
    \textbf{IO complexity breakdown for various methods.}
    $\cO(1)$ denotes the IO cost for a single data in the tensor across all layers and heads, which is equivalent to $\#heads*\#layer*dtype\_size$. The best among all methods in the table is in \textcolor{red}{red}, while the (potential) worst is in \textcolor{blue}{blue}. Query IO is omitted as it is $\cO(kl_n d_{head})$ for all methods. Here, $k$ is the number of QKV groups: for \algopt-Node $k=\#node$;for \algopt-Node-Chunk $k=\sum_{i=1}^{\#node} ceil(n_{i}/b_s)$, which is the node number after chunk wise;  for \algopt-Flatten, $k=N{tree}/b_s$, where $b_s$ is the block size of KV; for others, $k=1$. M in Tree Attention-M is short for Medusa~\citep{cai2024medusa}, while S in Tree Attention-S is short for SpecInfer~\citep{miao2023specinfer}.\label{tab:io_breakdown}
    }

    \begin{footnotesize}
        \resizebox{1.0\columnwidth}{!}{
            \begin{tabular}{ccccccc}
                \toprule
                \multicolumn{1}{c}{\textbf{Method}} & \multicolumn{1}{c}{\textbf{KV cache}}                   & \multicolumn{1}{c}{\bf{$\mQ \mK^\top$}} & \multicolumn{1}{c}{\bf{$\frac{\mQ \mK^\top}{s_{c}}$}} & \multicolumn{1}{c}{\bf{Mask(M)}}      & \multicolumn{1}{c}{\textbf{$\mM+\frac{\mQ \mK^\top}{s_{c}}$}} & \multicolumn{1}{c}{\bf{$\text{Softmax}$}} \\
                \midrule
                Naive Attention                     & \textcolor{blue}{$\cO(2d_{head} \sum_{i=1}^{l_n} N_i)$} & $\cO(2\sum_{i=1}^{l_n} N_i)$            & $\cO(2\sum_{i=1}^{l_n} N_i)$                          & \textcolor{red}{0}                    & \textcolor{red}{0}                                            & $\cO(2\sum_{i=1}^{l_n} N_i)$              \\
                \flashdecoding                      & \textcolor{blue}{$\cO(2d_{head} \sum_{i=1}^{l_n} N_i)$} & \textcolor{red}{0}                      & \textcolor{red}{0}                                    & \textcolor{red}{0}                    & \textcolor{red}{0}                                            & \textcolor{red}{0}
                \\
                Radix-Attention                     & \textcolor{blue}{$\cO(2d_{head} \sum_{i=1}^{l_n} N_i)$} & \textcolor{red}{0}                      & \textcolor{red}{0}                                    & \textcolor{red}{0}                    & \textcolor{red}{0}                                            & \textcolor{red}{0}                                              \\
                Tree Attention-M                    & \textcolor{red}{$\cO(2d_{head}N_{tree})$}               & \textcolor{blue}{$\cO(2l_n N_{tree})$}  & \textcolor{blue}{$\cO(2l_n N_{tree})$}                & \textcolor{blue}{$\cO(l_n N_{tree})$} & \textcolor{blue}{$\cO(2l_n N_{tree})$}                        & \textcolor{blue}{$\cO(2l_n N_{tree})$}    \\
                Tree Attention-S                    & \textcolor{blue}{$\cO(2d_{head}N_{tree}l_n)$}           & \textcolor{red}{$0$}                    & \textcolor{red}{$0$}                                  & $\cO(l_n N_{tree}/64)$                & \textcolor{red}{$0$}                                          & \textcolor{red}{$0$}                      \\
                \algopt-Node                        & \textcolor{red}{$\cO(2d_{head}N_{tree})$}               & \textcolor{red}{0}                      & \textcolor{red}{0}                                    & \textcolor{red}{0}                    & \textcolor{red}{0}                                            & \textcolor{red}{0}                        \\
                \algopt-Node-Chunk                        & \textcolor{red}{$\cO(2d_{head}N_{tree})$}               & \textcolor{red}{0}                      & \textcolor{red}{0}                                    & \textcolor{red}{0}                    & \textcolor{red}{0}                                            & \textcolor{red}{0}                        \\
                \algopt-Flatten                     & \textcolor{red}{$\cO(2d_{head}N_{tree})$}               & \textcolor{red}{0}                      & \textcolor{red}{0}                                    & $\cO(N_{tree})$                       & \textcolor{red}{0}                                            & \textcolor{red}{0}                        \\
                \bottomrule
            \end{tabular}
        }

    \end{footnotesize}
\end{table}

\begin{remark}[KV IO in SpecInfer]
    Though similar to \algopt, SpecInfer~\citep{miao2023specinfer} also employs a fused kernel for tree attention. As shown in \autoref{fig:treeattn_IO}, SpecInfer adopts \emph{Q-Guided Grouping}. Therefore,
    no IO is sharing for \kvcache among queries in SpecInfer: instead, each query will load the entire \kvcache of the tree independently, bringing significant IOs of the \kvcache as in \autoref{tab:io_breakdown}.
\end{remark}

\begin{remark}[IO in Radix Attention]
    Radix Attention~\citep{zheng2023efficiently} is essentially an implementation of Flash-Decoding~\citep{dao2023flashdecoding} utilizing paged and tree-structured memory management. As a result, the IO behavior is identical to that of Flash-Decoding, as shown in \autoref{tab:io_breakdown}.
\end{remark}

\begin{remark}[Causal mask IO]
    \algopt-Node splits the decoding tree by nodes without the need for causal masks.
    For more balanced calculations among SMs in GPUs, \algopt-Flatten evenly splits the decoding tree into blocks, with minimal IO cost for masks inspired by SpecInfer.
    This design reduces the IO overhead of masks significantly compared to the dense mask design in Medusa, as shown in \autoref{tab:io_breakdown}.
    \looseness=-1
\end{remark}

\subsection{Discussion of Workloads Generation\label{A:exp}}
\begin{figure}[!ht]
    \begin{center}
        \includegraphics[width=1.\columnwidth,height=0.2\textheight]{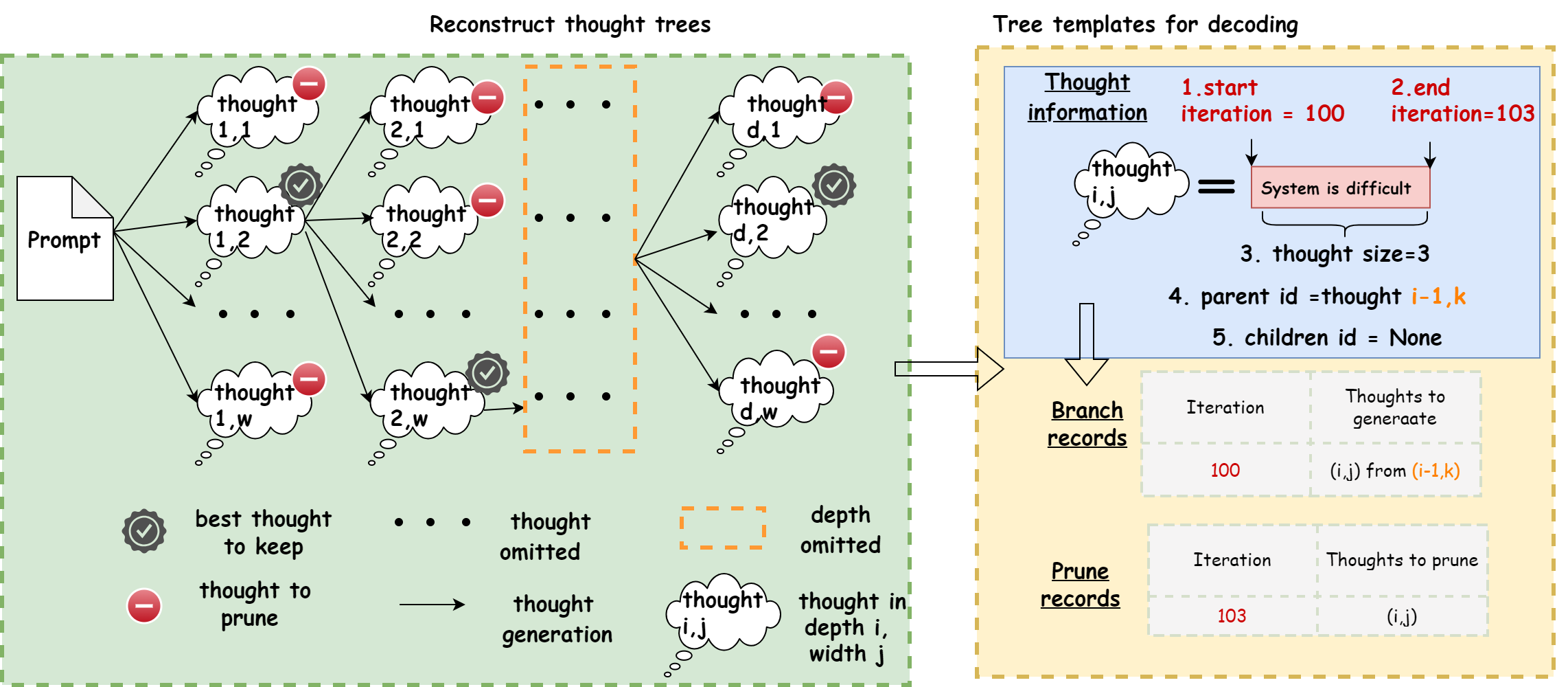}
    \end{center}
    \vspace{-1em}
    \caption{\small
        \textbf{The detailed procedure of reconstructing tree templates for multi-step reasoning}.
        (Left) Reconstructing reasoning trees from practical reasoning records as outlined in \citep{besta2023graph} involves capturing the following aspects:
        (1) the structure of trees, characterized by their depth $d$ and width $w$;
        (2) the token length associated with each thought;
        and (3) the best thought at each depth along with its corresponding score. For the task of document merging, the tree depth is set to $d=3$, with a width of $w=10$ at each depth.
        For sorting 128 numbers, the depth is reduced to $d=10$, while maintaining the same width of $w=10$. See details of tree topology for other multi-step reasoning tasks in \autoref{tab:workloadtree}.
        (Right) Utilizing the extracted thought information from Left, we can generate tree templates for decoding, encompassing \emph{branch records} and \emph{prune records}.
        These records are instrumental in guiding the tree decoding process to produce decoding trees that faithfully replicate the structure of the tree-of-thoughts.\label{fig:reconstruction}
    }

    \vspace{-1.em}
\end{figure}
\paragraph*{The rationality of workload settings.} To validate \algopt's acceleration across various decoding tree topologies, we compiled decoding trees from real tasks, covering the following three aspects:

\begin{itemize}[nosep, leftmargin=12pt]
    \item Few-shot prompting: This involves a two-level tree with a prompt prefix and multiple branches for suffix generation. As a case study, we fixed the prompt length at approximately 4000 tokens and varied the number of branches.
    \item  Multi-step reasoning~\citep{yao2023tree,hao2023reasoning,besta2023graph}: We recorded the tree shapes, prompts, and lengths of all thoughts from real reasoning task interactions~\citep{besta2023graph}, using these as guidance for tree decoding to validate \algopt's acceleration in thought generation of reasoning (the thought evaluation phase follows a similar pattern). See details of generation in \autoref{fig:reconstruction}.
    \item Speculative decoding~\citep{cai2024medusa,miao2023specinfer}: We used the token tree topology from Medusa~\citep{cai2024medusa} and recorded real interaction data with APPS~\citep{hendrycks2021measuring} as prompt dataset, including the length of accepted tokens at each step. This served as guidance to simulate the bottleneck of speculative decoding—the attention computation during the token verification phase.
\end{itemize}

\begin{table}[!ht]
    \setlength\tabcolsep{2pt}
    \centering
    \small
    \caption{\small
        \textbf{Details of generated workloads}.
        For multi-step reasoning, we include these 4 tasks from ~\cite{besta2023graph}: (1) Sorting 128 numbers (\textit{sorting} in short); (2) Document merging (\textit{document} in short); (3) Keyword counting (\textit{keyword} in short); (4) Set intersection (\textit{set} in short). $d$, and $w$ means depth and width of the tree, respectively. $t$ means the token tree size for speculative decoding, where the tree topology is from Medusa~\citep{cai2024medusa}.
        \label{tab:workloadtree}
    }
    \begin{footnotesize}
        \resizebox{1.0\columnwidth}{!}{
            \begin{tabular}{@{}cccc@{}}
                \toprule
                \textbf{Task}                         & \textbf{Tree Shape}                               & \textbf{Decoding Tree Source}                      & \textbf{Records Contents}                                                                             \\
                \midrule
                \multirow{4}{*}{Multi-step reasoning} & \multirow{1}{*}{\textit{sorting}: $d=10$, $w=10$} & \multirow{4}{*}{\makecell{ToT-BFS in \\ ~\cite{besta2023graph}}} & \multirow{4}{*}{\makecell{Prompt~\citep{besta2023graph},\\tree shape, thought size,\\ branch records,prune records}}  \\
                                                      & \textit{document}: $d=3$, $w=10$                  &                                                    &                                                                                                       \\
                                                      & \textit{keyword}:$d=5$, $w=10$                    &                                                    &                                                                                                       \\
                                                      & \textit{set}:$d=8$, $w=10$                        &                                                    &                                                                                                       \\\midrule
                Few-shot prompting                    & $d=1$, $w=10, 20,30$                              & --                                                 & --                                                                                                    \\\midrule
                Speculative decoding                  & $t=32,64,128,256$                                 & Medusa~\citep{cai2024medusa}                        & \makecell{APPS~\citep{hendrycks2021measuring} \\ Prompt, token tree shape, \\ accepted token length per step}
                \\
                \bottomrule
            \end{tabular}
        }
    \end{footnotesize}
    \vspace{-1em}
\end{table}

\paragraph*{The rationality of our experiment paradigm.} Our experimental paradigm involves: first, obtaining decoding trees from real tree-based decoding tasks, and second, replicating these decoding trees exactly within the same framework by enforcing LLM inference, to investigate the impact of attention acceleration on wall clock time performance. This paradigm has two advantages:
\begin{itemize}[nosep, leftmargin=12pt]
    \item We can utilize decoding trees from real tasks as a benchmark within a unified system, enabling fair comparison of different attention algorithms in terms of decoding latency. This comparison is possible despite the algorithms being based on distinct systems, such as variations in memory management implementations for their kernels.

    \item We consider both the unique characteristics of tasks with diverse tree structures and the broader applicability of general tree-based decoding. See details of generated workloads for other multi-step reasoning tasks in \autoref{tab:workloadtree}.
\end{itemize}

\subsection{Additional Results \label{A:AEXP}}

\paragraph{Microbench of \algopt-Node for GPU utilization.} We test the \algopt-Node and \algopt-Flatten on a speculative decoding with 64 queries and a prompt with 4k tokens. For \algopt-Node, the QKV partitioning is unbalanced as the node of the prompt is much longer than others (1 token for each query). As shown in \autoref{tab:microbench_gpu}, the metrics of GPU utilization are measured by NVIDIA Nsight Compute~\citep{nvidia_nsight_compute}. We can see \algopt-Flatten is better than \algopt-Node in both memory utilization (\emph{Memory Throughput Ratio}) and calculation utilization (\emph{Compute Throughput Ratio} and \emph{Low Utilization Time Ratio}).

\begin{table}[!h]
    \setlength\tabcolsep{2pt}
    \centering
    \small
    \caption{\small
    \textbf{[GPU Utilization Microbenchmark] Latency of a single layer of Attention (in $\mu$s), SM Compute Throughput Ratio, Memory Throughput Ratio, and Low Utilization Ratio for \algopt on an NVIDIA A100 (80GB) using the LLama3-8B model (GQA).} The workload is speculative decoding with 64 queries and a prompt with 4k tokens. The \emph{Compute Throughput Ratio} refers to the utilization of the Streaming Multiprocessors (SMs) in the GPU. The \emph{Memory Throughput Ratio} represents the ratio between actual memory throughput and maximum bandwidth. The \emph{Low Utilization Time Ratio} is defined as the proportion of time when the \emph{Compute Throughput Ratio} falls below 5$\%$.
    \label{tab:microbench_gpu}
    }
    \begin{footnotesize}
        \resizebox{0.9\columnwidth}{!}{
            \begin{tabular}{c|c|c|c|c}
                \toprule
                 Method  & Attention Latency                                   & Compute Throughout Ratio
                 & Memory Throughout Ratio
                 & Low Utilization Time Ratio
                                               
                \\ \midrule

                                        \algopt-Node      &           	961.38                     &      	7.60$\%$           & 	17.39$\%$  & 82.35$\%$
                \\
                                        \algopt-Flatten          &           226.82                        &      	21.19$\%$           & 51.91$\%$ & 0.00$\%$

                \\
                \bottomrule
            \end{tabular}
        }

    \end{footnotesize}
\end{table}

\paragraph*{Attention latency and IOs with breakdowns.} The details of attention latency and IO comparison among \algopt and baselines are in \autoref{tab:e2e_comp} and \autoref{tab:PIO_comp}, respectively.  Note that the attention latency does not include the memory management overheads, which is the bottleneck for \treeAttn. For example, it takes more than $80\%$ of end to end latency in speculative decoding with 32 queries. 

\begin{table}[h]
    \centering
    \caption{\textbf{Inference accuracy of \algopt in attention score and perplexity (PPL)}. PPL is calculated after 400 iterations of decoding. Vanilla Attention is the implementation from  Huggingface Transformers.\label{tab:infer_acc} }
    \begin{footnotesize}
    \resizebox{0.7\columnwidth}{!}{
    \begin{tabular}{lcc|cc|cc}
    &  \multicolumn{2}{c}{Relative Attention Error ($\downarrow$)} & \multicolumn{2}{c}{Perplexity (PPL) ($\downarrow$)} & \multicolumn{2}{c}{Relative PPL Error ($\downarrow$)} \\
    \toprule
    \multirow{2}{*}{Attention Method} & \multicolumn{6}{c}{Attention Variations} \\ 
    \cmidrule(lr){2-7}
     & MHA & GQA  & MHA & GQA   & MHA & GQA  \\
    \midrule
    
    Vanilla Attention & - & - & 1.000 & 1.002 & - & - \\
    Radix Attention & 0.545\% & 0.540\% & 1.000 & 1.002 & $1 \times 10^{-6}$ & $1 \times 10^{-6}$ \\
    \algopt-Node-Chunk & 0.403\% & 0.403\% & 1.000 & 1.002 & $1 \times 10^{-6}$ & $4 \times 10^{-6}$ \\
    \algopt-Flatten & 0.407\% & 0.404\% & 1.000 & 1.002 & $1 \times 10^{-6}$ & $9 \times 10^{-7}$ \\
    \bottomrule
    \end{tabular}
    }
    
    \end{footnotesize}
    \vspace{-5mm}
    \end{table}

\paragraph{Inference accuracy of \algopt-Node-Chunk and \algopt-Flatten.} In Section \ref{subsec:Preliminary}, equation \ref{eq:seg_softmax} shows \algopt (including \algopt-Node, \algopt-Node-Chunk and \algopt-Flatten ) and vanilla attention are mathematically equivalent, which means \algopt is accurate. As demonstrated in \autoref{tab:infer_acc}, the \algopt attention scores may slightly differ (around 0.4$\%$ relative error) compared to vanilla attention in Huggingface Transformers, but the generated tokens will hardly be different as well as PPL (1e-6 relative PPL error in the right part of \autoref{tab:infer_acc}). This discrepancy arises because floating-point operations on GPUs do not adhere to the associative law, even when two calculation processes are mathematically equivalent. Similar issues occur in other methods like radix attention and Flash-Decoding that introduce online Softmax and reduction as well, resulting in approximately 0.5$\%$ relative attention score errors.

\begin{table}[!tbp]
    \setlength\tabcolsep{2pt}
    \centering
    \small
    \caption{\small
    \textbf{Average attention latency (in seconds) for tree-based decoding and its impact on decoding latency.}
    Here, $b$ represents the tree width, and $t$ denotes the token tree size (i.e., the number of tree-structured queries).
    \textit{Attention Speedup over the best attention} refers to the speedup of \algopt-Flatten compared to the best baseline (typically \textit{\treeAttn}) in attention calculation. \textcolor{blue}{\textit{Radix Attention}} is the best baseline for decoding latency. Note that KV cache management is not included in the attention latency.
    $\star$ denotes out-of-memory (OOM) errors for the A100 80GB GPU.
    For more details on decoding latency, see \autoref{tab:e2e_comp}.
    \label{tab:e2eAttn_comp}}
    
    \begin{footnotesize}
        \resizebox{0.95\columnwidth}{!}{
            \begin{tabular}{ccccccccccccc}
                \toprule
                \multirow{2}{*}{Memory} & \multirow{2}{*}{Method}                                      & \multicolumn{3}{c}{Few-shot Prompting }
                                        & \multicolumn{4}{c}{Multi-Step Reasoning}
                                        & \multicolumn{4}{c}{Speculative Decoding}
                \\ \cmidrule(lr){3-5} \cmidrule(lr){6-9} \cmidrule(lr){10-13}
                                        &                                                              & \texttt{b=20}                           & \texttt{b=30}          & \texttt{b=50}
                                        & \textit{Sorting}                                             & \textit{Document}                       & \textit{Keyword}       & \textit{Set}
                                        & \texttt{t=32}
                                        & \texttt{t=64}                                                & \texttt{t=128}                          & \texttt{t=256}
                \\ \midrule
                \multirow{2}{*}{Unpaged}
                                        & \flashdecoding                                               & 43.49                                   & 66.10                  & 110.09
                                        & 160.67                                                       & 105.80                                  & 12.14                  & 19.96
                                        & 340.09                                                       & 692.88                                  & $\star$                & $\star$

                \\
                                        & \treeAttn                                                    & \textcolor{red}{3.93}                   & 7.51                   & \textcolor{red}{9.57}
                                        & \textcolor{red}{38.64}                                       & 29.10                                   & \textcolor{red}{2.62}  & \textcolor{red}{3.96}
                                        & \textcolor{red}{22.40}                                       & \textcolor{red}{26.31}                  & \textcolor{red}{41.10} & \textcolor{red}{68.28}

                \\

                \midrule
                \multirow{3}{*}{Paged}
                                        & \textcolor{blue}{Radix Attention}                            & 5.99                                    & \textcolor{red}{7.30}  & 9.96
                                        & 39.37                                                        & \textcolor{red}{24.69}                  & 3.11                   & 5.13
                                        & 25.73                                                        & 40.47                                   & 76.10                 & 145.43
                \\
                                        & \algopt-Flatten                          .                   & 3.47                                    & 4.07                   & 5.87
                                        & 28.41                                                        & 21.45                                   & 2.57                   & 3.83
                                        & 13.15                                                        & 16.79                                   & 24.46                  & 40.56
                \\
                \midrule
                                        & Attention Speedup over  \textcolor{red}{the best attention}. & 1.13$\times$                            & 1.63$\times$           & 1.70$\times$
                                        & 1.36$\times$                                                 & 1.15$\times$                            & 1.02$\times$           & 1.03$\times$
                                        & 1.70$\times$                                                 & $1.57\times$                            & $1.68\times$           & $1.68\times$
                \\   \midrule                      &\textit{Attention Speedup over \textcolor{blue}{Radix Attention} }           & 1.73$\times$                                 & 1.63$\times$                 & 1.70$\times$
                                        & 1.39$\times$                                                 & 1.15$\times$                            & 1.21$\times$           & 1.34$\times$
                                        & 1.96$\times$                                                 & $2.41\times$                            & $3.11\times$           & $3.59\times$
                \\   \midrule                      &\textit{Decoding Speedup over \textcolor{blue}{Radix Attention} }           & 1.24$\times$                                &  1.28$\times$                & 1.33$\times$
                                        & 1.10$\times$                                                 & 1.03$\times$                            & 1.03$\times$           & 1.05$\times$
                                         & 1.29$\times$                                                           & $1.50\times$                            & $1.91\times$       & $2.23\times$

                \\

                \bottomrule
            \end{tabular}
        }

        \vspace{-1em}
    \end{footnotesize}
\end{table}

\begin{table}[!tbp]
    \setlength\tabcolsep{2pt}
    \centering
    \small
    \caption{\small
    \textbf{Average end-to-end IO (TB) during decoding. } Data format is Left/Right: \textit{(Left)} KV Cache IO; \textit{(Right)} partial results IO, including $\mQ \mK^T$,$\mQ \mK^\top/{s_{c}}$,  Mask $M$, $\mM+\mQ \mK^\top/{s_{c}}$ and $\text{Softmax}$.  $b$ means tree width. $t$ denotes the token tree size (i.e., the number of tree-structured queries).$\star$ means out of memory for A100 80GB.
    \label{tab:PIO_comp}}
    \begin{footnotesize}
        \resizebox{1.00\columnwidth}{!}{
            \begin{tabular}{ccccccccccccc}
                \toprule
                \multirow{2}{*}{Method} & \multicolumn{3}{c}{Few-shot Prompting }
                                        & \multicolumn{4}{c}{Multi-Step Reasoning}
                                        & \multicolumn{3}{c}{Speculative Decoding}
                \\ \cmidrule(lr){2-4} \cmidrule(lr){5-8} \cmidrule(lr){9-12}
                                        & \texttt{b=20}                            & \texttt{b=30}
                                        & \texttt{b=50}
                                        & \textit{Sorting}                         & \textit{Document} & \textit{Keyword} & \textit{Set} & \texttt{t=32 } & \texttt{t=64 } & \texttt{t=128}
                                        & \texttt{t=256}                                                                                                                                    \\ \midrule

                \flashdecoding          & 17.62/0.00                               & 26.43/0.00        & 44.05/0.00
                                        & 59.96/0.00                               & 39.74/0.00        & 4.68/0.00        & 7.01/0.00
                                        & 128.72/0.00                              & 255.16/0.00       & $\star$          & $\star$                                                         \\
                \treeAttn               & 1.68/1.05                                & 2.10/1.98         & 2.94/4.61
                                        & 12.40/3.69                               & 10.57/3.24        & 0.58/0.18        & 1.04/0.27
                                        & 4.02/4.03                                & 4.15/8.33         & 4.18/16.77       & 4.32/34.70
                \\
                Radix Attention         & 17.62/0.00                               & 26.43/0.00        & 44.05/0.00
                                        & 59.96/0.00                               & 39.74/0.00        & 4.68/0.00        & 7.01/0.00
                                        & 131.45/0.00                              & 256.79/0.00       & 522.05/0.00      & 1044.10/0.00
                \\
                \algopt-Flatten         & 1.68/0.00                                & 2.10/0.00         & 2.94/0.00
                                        & 12.40/0.01                               & 10.57/0.01        & 0.58/0.00        & 1.04/0.00
                                        & 4.10/0.00                                & 4.11/0.00         & 4.16/0.00        & 4.35/0.00
                \\ \midrule
                IO reduction of \algopt-Flatten($\%$)
                                        & 90.47/100.00                             & 92.1/100.00       & 93.33/100.00
                                        & 79.32/99.73                              & 73.40/99.70       & 87.61/100.00     & 85.16/100.00
                                        & 96.88/100.00                             & 98.40/100.00      & 99.20100.00      & 99.58/100.00                                                    \\
                \bottomrule
            \end{tabular}
        }

    \end{footnotesize}
\end{table}

\begin{figure}[!h]
    \centering
    \includegraphics[width=0.8\linewidth]{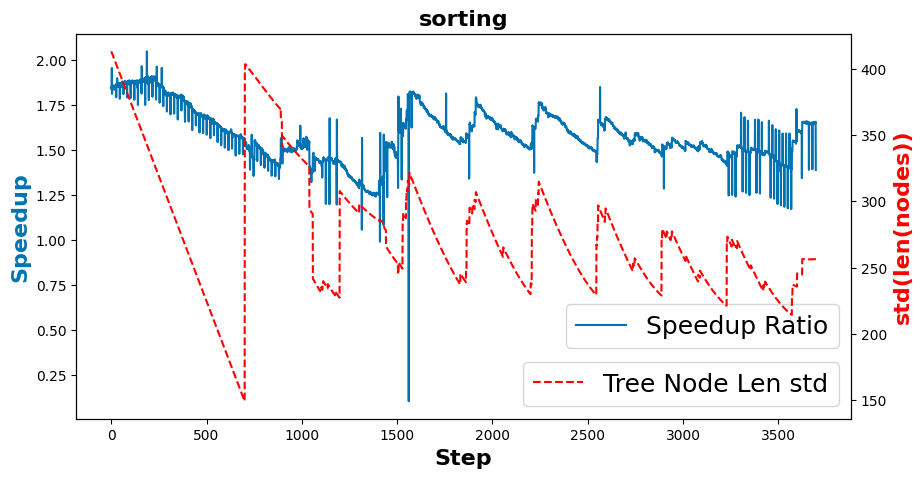}
    \caption{
        Comparison of split strategies \algopt-Node and \algopt-Flatten in \textit{sorting} task.
        \textit{Speedup ratio} refers to the ratio between the per iteration latency of \algopt-Node and \algopt-Flatten.
        \textit{Tree Node Len std} represents the standard deviation of the tree node lengths for each iteration.
    }
    \label{SC_sorting}
\end{figure}

\paragraph{Dynamic behaviors: per iteration latency.} We visualize the per-iteration latency of \algopt-Node and \algopt-Flatten for a tree in the multi-step reasoning task--\textit{sorting}  in \autoref{SC_sorting}, as the size and topology of the decoding tree change with each iteration. This comparison highlights the sensitivity of these two split strategies to changes in tree size. We observe a strong positive correlation between the ratio of per-iteration latency of \algopt-Node and \algopt-Flatten (Speedup Ratio) and the dispersion of tree node sizes. This correlation arises because the performance of \algopt-Flatten remains relatively stable, whereas the performance of \algopt-Node is more strongly influenced by the topology of the tree. \algopt-Flatten provides a stable speedup of approximately 1.75$\times$ compared to \algopt-Node.

\begin{figure}[!ht]
    \centering
    \subfloat[\small
        Tree width is 10.
        \looseness=-1
        \label{F10}
    ]{\includegraphics[width=0.48\linewidth]{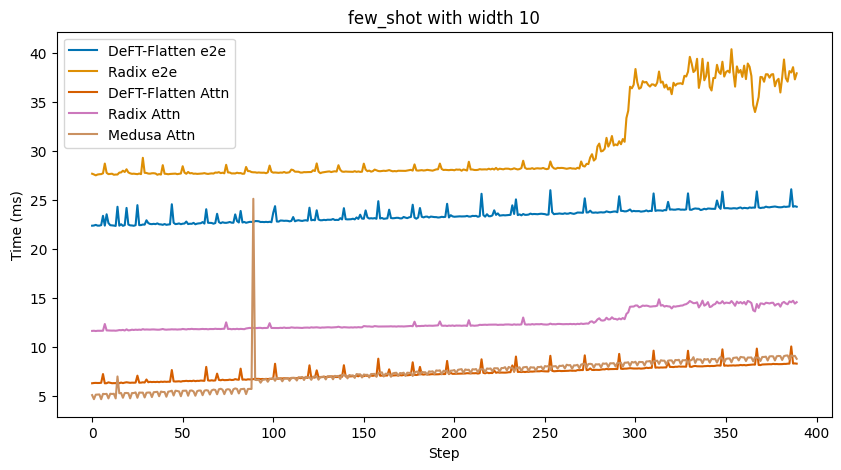}}
    \hfill
    \subfloat[\small
        Tree width is 20.
        \looseness=-1
        \label{F20}
    ]{\includegraphics[width=0.48\linewidth]{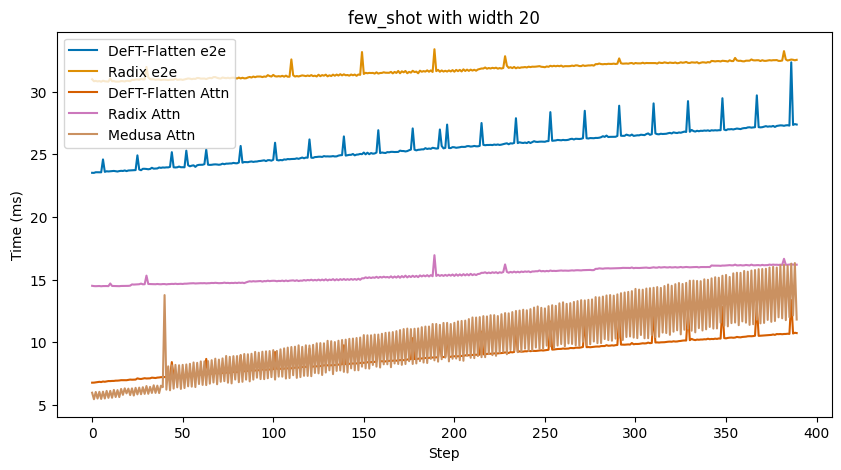}}
    \vspace{-0.5em}
    \subfloat[\small
        Tree width is 30..
        \looseness=-1
        \label{F30}
    ]{\includegraphics[width=0.48\linewidth]{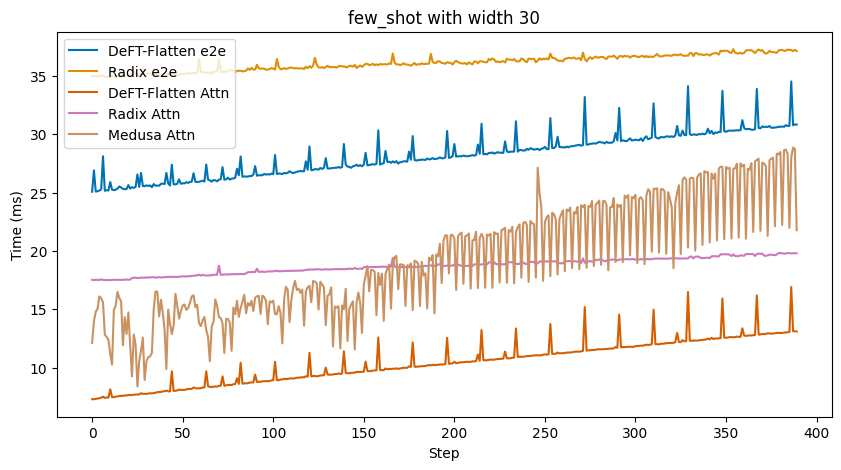}}
    \hfill
    \subfloat[\small
        Tree width is 50.
        \looseness=-1
        \label{F50}
    ]{\includegraphics[width=0.48\linewidth]{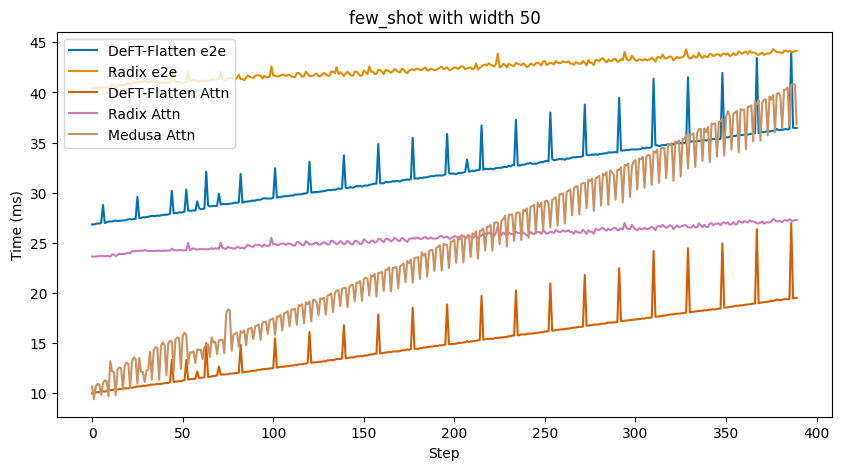}}
    \vspace{-0.8em}
    \caption{ \small
        Per iteration latency for few-shot prompting tasks with different tree width. \textit{e2e} means decoding latency(optimal end-to-end latency), while \textit{Attn} means only the attention overhead. \label{PL_fewshot}
        \looseness=-1
    }
    \vspace{-1em}
\end{figure}

\paragraph*{Ablation: The influence of width in decoding trees.} We observe that the effectiveness of attention speedup varies with different decoding tree topologies. Considering the simplest tree structure, a prompt with several suffixes—given a prompt that is not very short, one of the most important factors for speedup is the extent to which we can reuse its KV cache IO. This can be measured by the width of the tree. More specifically, it is determined by the number of queries per iteration. Therefore, we fix the prompt length at 4000 and vary the width of the decoding tree in few-shot prompting (which also indicates how many requests share the same prompt). Then, as shown in \autoref{PL_fewshot}, we evaluate \algopt-Flatten with the best baseline in attention calculation-- \treeAttn~\citep{cai2024medusa} (Medusa-Attn in the figure), as well as the best baseline in decoding latency-- Radix Attention~\citep{zheng2023efficiently}, for the per-iteration latency over time. We have the following observations:
\begin{enumerate}[nosep, leftmargin=12pt]
    \item When the tree width is 10, the attention overhead of \algopt-Flatten is nearly the same as \treeAttn because the IO overhead of the dense causal mask (DCM) is small compared to that of the KV cache, but it is still 2$\times$ faster in attention latency than Radix Attention thanks to the KV IO reuse.
    \item As the tree width increases, the attention computation overhead of \treeAttn grows faster because the size of the DCM is directly related to the tree width. A larger tree width means the IO of the DCM grows rapidly.
    \item Since the tree topology consists of a fixed prefix with several suffixes, a larger tree width allows the prompt prefix's KV cache to be reused more frequently during IO. This leads to a more significant decoding speedup—\(1.24\times\) with a width of \(w=20\), and \(1.33\times\) with a width of \(w=50\)—compared to Radix Attention.
    \item As iterations progress, the length of the suffixes gradually approaches the length of the prefix, leading to a decrease in the speedup of \algopt-Flatten compared with Radix Attention.
\end{enumerate}

\begin{figure}[h]
    \begin{center}
        \includegraphics[height=0.2\textheight]{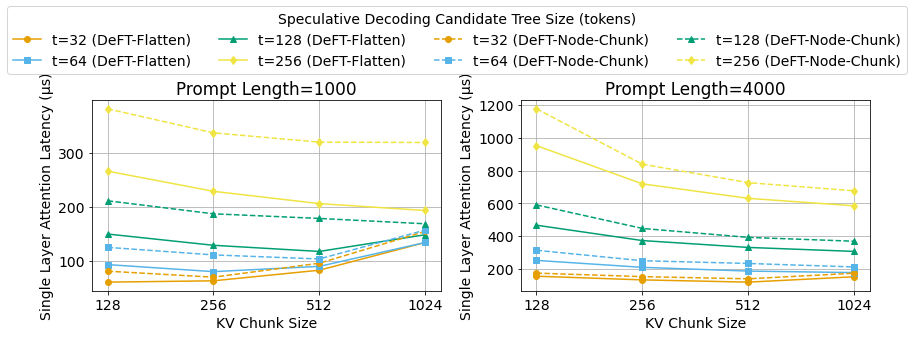}
    \end{center}
    \vspace{-1.em}
    \caption{\small
        \textbf{Ablation study for KV chunk size with \algopt}. $t$ is the token tree size in speculative decoding. \label{fig:ablation_chunk}}
    \vspace{-1.em}
\end{figure}

\paragraph{Ablation: The influence of chunk size in KV splitting.} In the implementation of \algopt-Flatten and \algopt-Node-Chunk, we selected a regular size (128) in General matrix multiply (GEMM) as the block/chunk size of KV cache during the attention calculation.
We added an ablation study of chunk size influence on speedup, as shown in \autoref{fig:ablation_chunk}. The chunk size selection is a trade-off between IO redundancy and threadblock scheduling: a larger chunk size means less redundancy of Query IO but may cause potential idle SMs of GPUs due to fewer threadblocks during GPU scheduling.
Conclusions of \autoref{fig:ablation_chunk}: (1) The best chunk size is influenced by both sequence length and query numbers (batch size); (2) DeFT-Flatten can outperform DeFT-Node-Chunk in all chunk sizes we test.

\begin{figure}[!ht]
    \centering
    \includegraphics[width=0.95\linewidth]{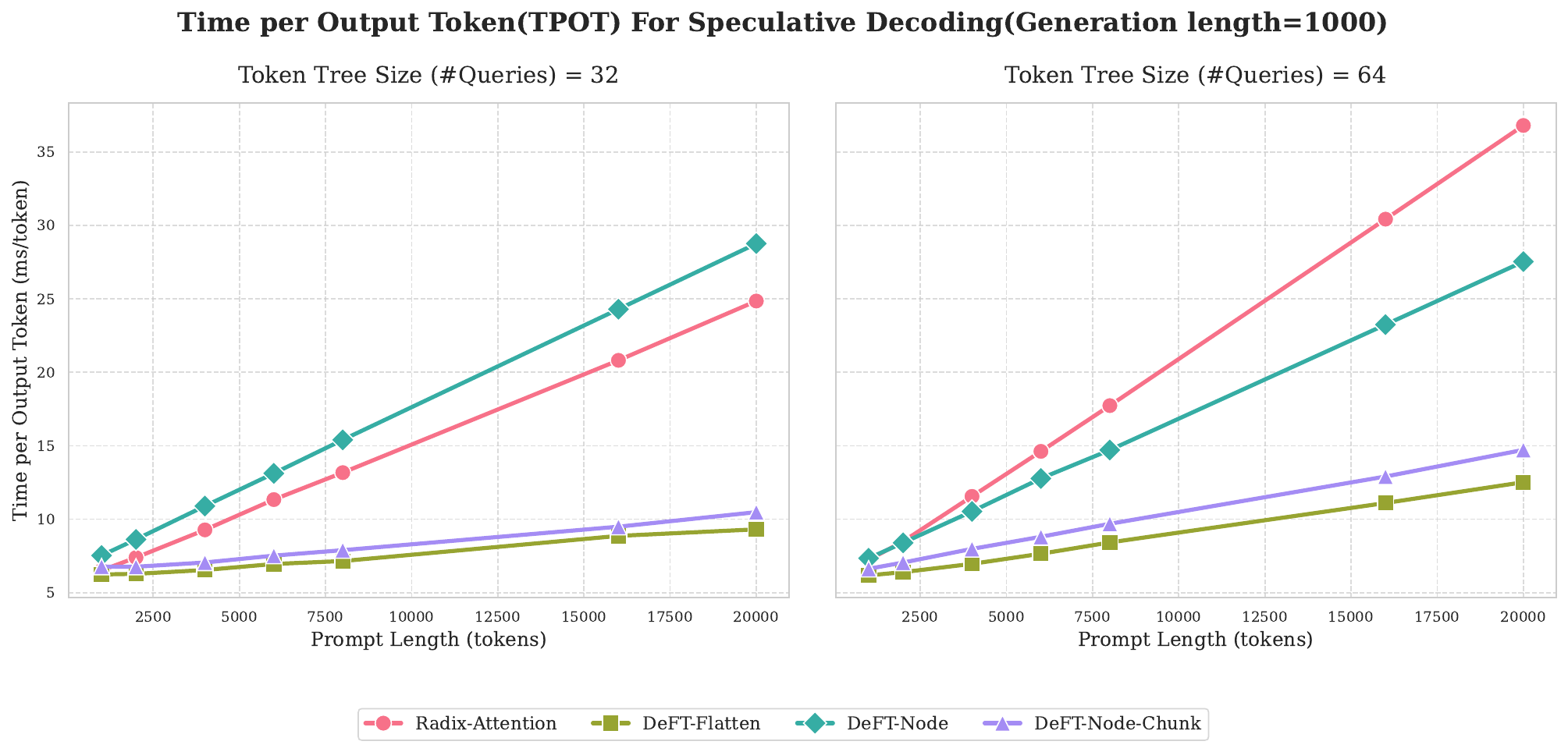}
    \caption{
        Time per output token(TPOT) of \algopt with different prompt lengths in speculative decoding.}
    \label{abla:tpot}
\end{figure}

\begin{figure}[!ht]
    \centering
    \includegraphics[width=0.95\linewidth]{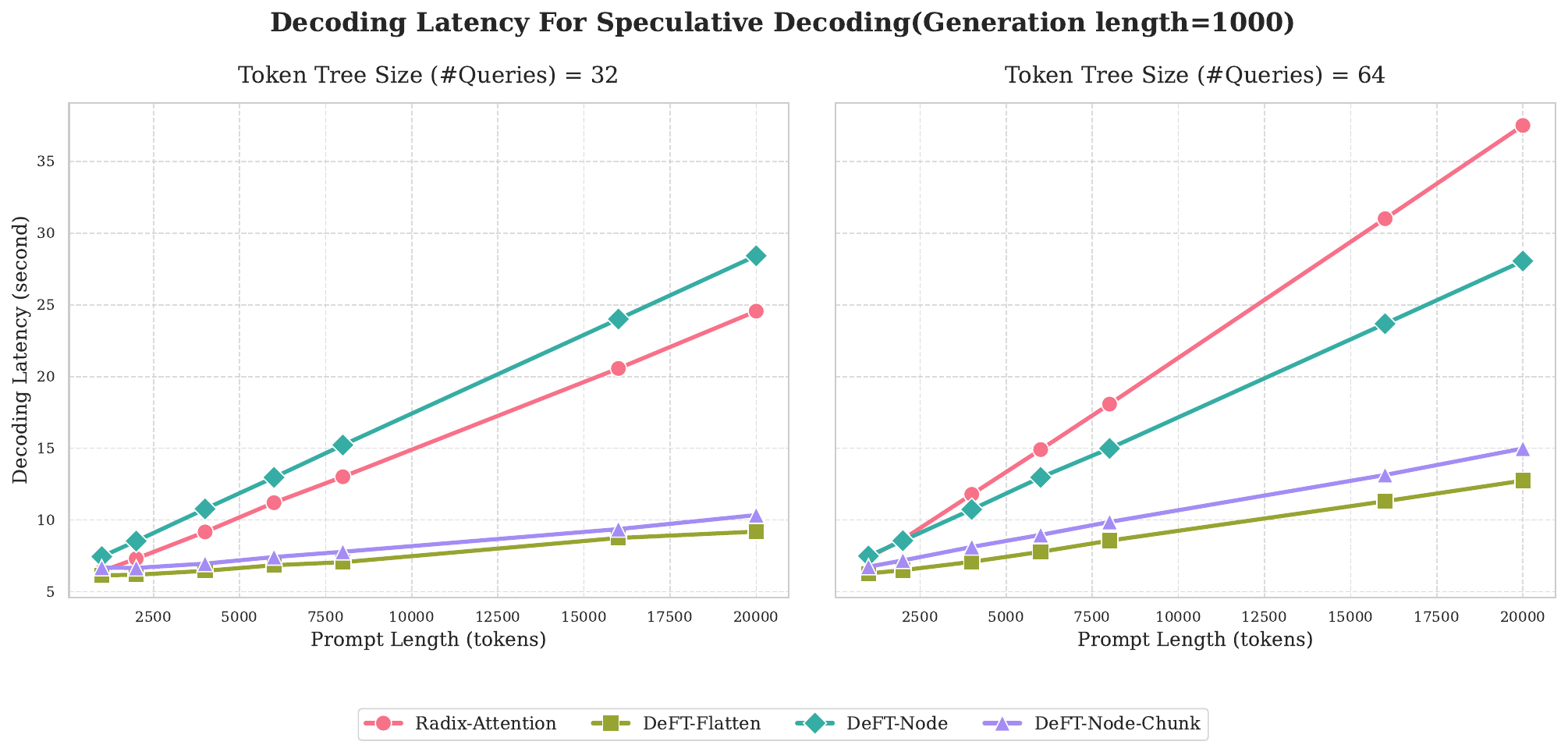}
    \caption{
        Decoding latency of \algopt with different prompt lengths in speculative decoding.}
    \label{abla:decoding_latency}
\end{figure}

\begin{figure}[!ht]
    \centering
    \includegraphics[width=0.95\linewidth]{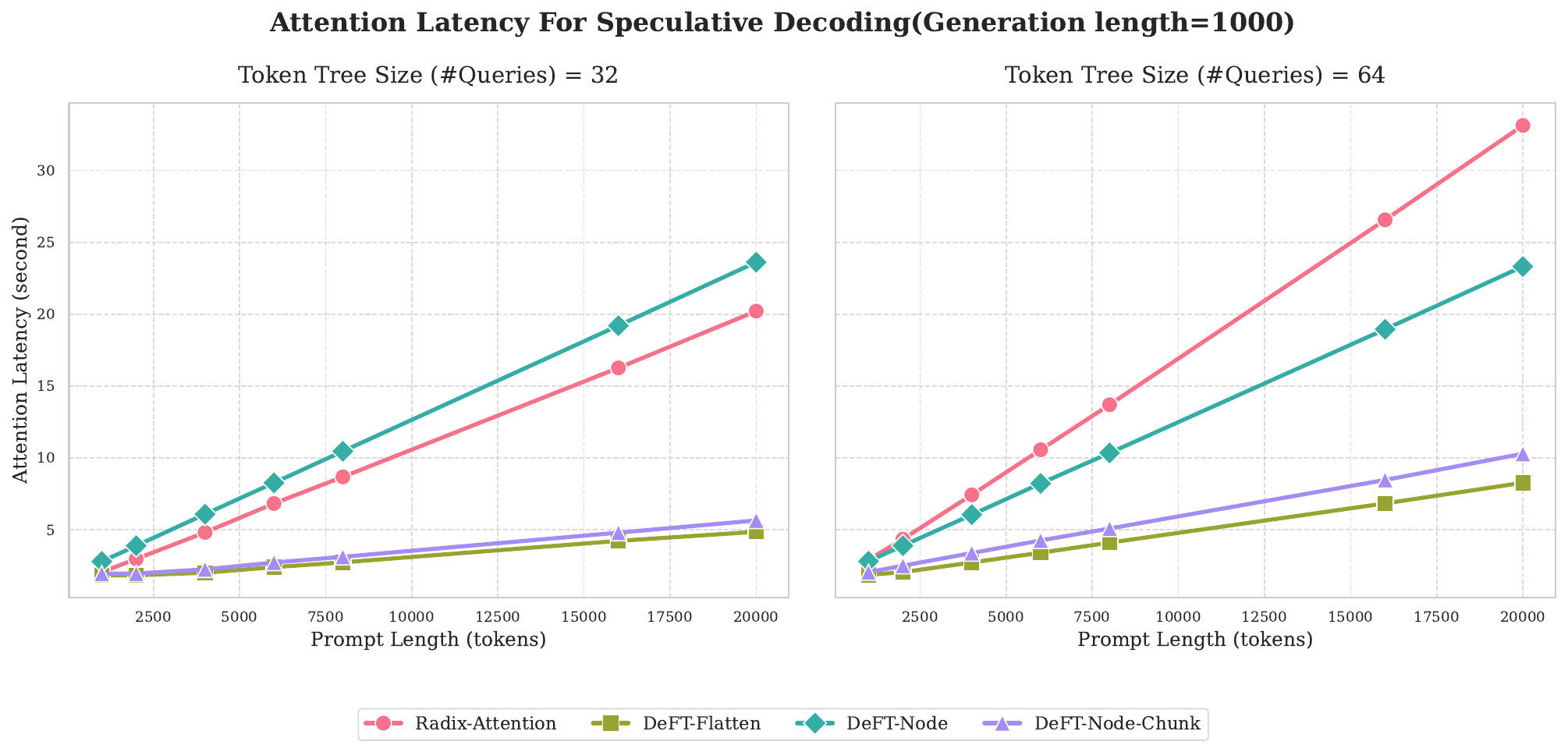}
    \caption{
        Attention latency of \algopt with different prompt lengths in speculative decoding.}
    \label{abla:attn_latency}
\end{figure}

\paragraph{Ablation: The influence of prompt length.} We set the prompt length from 1k to 20k tokens, and the generation length is set to 1k tokens, to simulate a long prompt short generation case. Our workloads are speculative decoding with token tree size of 32 and 64. The ablation is based on Llama3.1-8B model and NVIDIA A100 80GB. We have three metrics: time per output token (\autoref{abla:tpot}), decoding latency (\autoref{abla:decoding_latency}) and attention latency (\autoref{abla:attn_latency}).

\paragraph{Ablation: The influence of both model size and prompt length.} In \autoref{tab:e2e_chunk} and \autoref{tab:model_size1} of Section \ref{subsec:ablation}, we show the influence of prompt length and model size, individually. Here, we present the ablation study of both model size and prompt length for \textit{sorting} task, as shown in \autoref{tab:ablation_model_prompt}. With a longer prompt, DeFT shows more pronounced speedup in the same model, since the attention overhead is proportional to the token count in the decoding tree, while the FFN overhead remains nearly constant for the same model. For a fixed prompt length, with the larger model—Codellama-34B, \algopt-Flatten achieves slightly reduced but still significant (up to 1.28x) decoding speedup. The performance reduction is attributed to a lower A/F-LR, as the FFN overhead is greater in larger models.

\begin{table}[!h]
    \setlength\tabcolsep{2pt}
    \centering
    \small
    \caption{\small
        \textbf{ [Ablation Study of Model Size and Prompt Length] Comparison of decoding speedup and Attention/FFN latency ratio (\textcolor{red}{A/F-LR}) between \algopt and \textcolor{blue}{Radix Attention} for Codellama-34B and Codellama-7B across varying prompt lengths in the \texttt{sorting} reasoning task.}  \textcolor{blue}{Radix Attention} is the best baseline in decoding latency. 
        \label{tab:ablation_model_prompt}
    }
    \begin{footnotesize}
        \resizebox{1.\columnwidth}{!}{
            \begin{tabular}{cc|c|c|c}
                \toprule
                \multirow{1}{*}{Metric} & Model Size    & Prompt Length\texttt{=1k} 
                                        & Prompt Length\texttt{=5k} 
                                        & Prompt Length\texttt{=8k}
               
                \\
                \midrule
                \multirow{2}{*}{Decoding latency Speedup}
                & 7B & 1.09$\times$ & 1.37$\times$ & 1.53$\times$ \\
                & 34B & 1.03$\times$ & 1.18$\times$ & 1.28$\times$
               
                \\ \midrule    
                \multirow{2}{*}{Radix Attention's \textcolor{red}{A/F-LR}}
                & 7B & 1.12  & 1.89 & 2.50  \\
                & 34B & 0.48 & 0.86 & 1.16

                \\ \midrule    
                \multirow{2}{*}{\algopt-Flatten's \textcolor{red}{A/F-LR}}
                & 7B & 0.89  & 1.09 & 1.25  \\
                & 34B & 0.42 & 0.57 & 0.67

                \\
                \bottomrule
            \end{tabular}
        }

    \end{footnotesize}

\end{table}

\paragraph{Ablation: Different GPUs.} See \autoref{tab:e2eAttn_4090}. \algopt-Flatten can have obvious speedup in RTX 4090 as well because the memory hierarchy of GPUs is nearly the same— large but slow global memory and small but fast shared memory.

\begin{table}[!h]
    \setlength\tabcolsep{2pt}
    \centering
    \small
    \caption{\small
        \textbf{ [Different GPUs] Speedup of \algopt in average attention latency (second) with NVIDIA RTX 4090 (24GB) for LLama3-8B model(GQA).}  \textcolor{blue}{Radix Attention} is the best baseline in decoding latency. 
        \label{tab:e2eAttn_4090}
    }
    \begin{footnotesize}
        \resizebox{1.\columnwidth}{!}{
            \begin{tabular}{ccccc}
                \toprule
                \multirow{1}{*}{Memory} & \multirow{1}{*}{Method}                                      & \multicolumn{1}{c}{Few-shot Prompting-\texttt{b=30}}
                                        & \multicolumn{1}{c}{Multi-Step Reasoning-\textit{Sorting}}
                                        & \multicolumn{1}{c}{Speculative Decoding-\texttt{t=64} }
                                        
                \\ \midrule
                                       
                \multirow{3}{*}{Paged}
                                        & \textcolor{blue}{Radix Attention}                           &      4.26                              &      26.36           & 	33.63
                \\
                                        & \algopt-Node-Chunk      &           	3.07                         &      	24.61           & 	15.39
                \\
                                        & \algopt-Flatten          &           2.95                         &      	23.86           & 14.04

                \\ \midrule    
                &\textit{Attention Speedup over \textcolor{blue}{the best decoding} }           & 1.44$\times$                                &  1.10$\times$                & 2.40$\times$

                \\
                \bottomrule
            \end{tabular}
        }

    \end{footnotesize}
\end{table}

\paragraph{Ablation: Different Model Architectures.} See \autoref{tab:e2eAttn_34b} and \autoref{tab:e2eAttn_7b}. \algopt-Flatten can both accelerate the attention computation of LLM models with different architectures (MHA and GQA) significantly.

\begin{table}[!h]
    \setlength\tabcolsep{2pt}
    \centering
    \small
    \caption{\small
        \textbf{ [Different Model Architectures(GQA)] Speedup of \algopt in average attention latency (second) with NVIDIA A100(80GB) for Codellama-34B model(GQA).}  \textcolor{blue}{Radix Attention} is the best baseline in decoding latency. 
        \label{tab:e2eAttn_34b}
    }
    \begin{footnotesize}
        \resizebox{1.\columnwidth}{!}{
            \begin{tabular}{ccccc}
                \toprule
                \multirow{1}{*}{Memory} & \multirow{1}{*}{Method}                                      & \multicolumn{1}{c}{Few-shot Prompting-\texttt{b=30} }
                                        & \multicolumn{1}{c}{Multi-Step Reasoning-\textit{Sorting}  }
                                        & \multicolumn{1}{c}{Speculative Decoding-\texttt{t=64} }

                \\
                \midrule
                \multirow{3}{*}{Paged}
                                        & \textcolor{blue}{Radix Attention}          &       16.85                                &  95.14                 
                                        &  164.33 
                \\
                                        & \algopt-Node-Chunk
                                        &    16.15                         
                                        &   103.15                 
                                        & 81.74  
                \\
                                        & \algopt-Flatten                            &     9.62                               
                                        &       84.30             
                                        &     48.76

                \\ \midrule    
                &\textit{Attention Speedup over \textcolor{blue}{the best decoding} }          
                & 1.75$\times$                                
                &  1.13 $\times$                
                & 3.37$\times$
                
                \\
                \bottomrule
            \end{tabular}
        }

    \end{footnotesize}

\end{table}

\begin{table}[!h]
    \setlength\tabcolsep{2pt}
    \centering
    \small
    \caption{\small
        \textbf{ [Different Model Architectures(MHA)] Speedup of \algopt in average attention latency (second) with NVIDIA A100(80GB) for Codellama-7B model(MHA).}  \textcolor{blue}{Radix Attention} is the best baseline in decoding latency. 
        \label{tab:e2eAttn_7b}
    }
    \begin{footnotesize}
        \resizebox{1.\columnwidth}{!}{
            \begin{tabular}{ccccc}
                \toprule
                \multirow{1}{*}{Memory} & \multirow{1}{*}{Method}                                      & \multicolumn{1}{c}{Few-shot Prompting-\texttt{b=30} }
                                        & \multicolumn{1}{c}{Multi-Step Reasoning-\textit{Sorting}  }
                                        & \multicolumn{1}{c}{Speculative Decoding-\texttt{t=64} }

                \\
                \midrule
                \multirow{3}{*}{Paged}
                                        & \textcolor{blue}{Radix Attention}                           &       12.39                                &        53.96           &  96.55
                \\
                                        & \algopt-Node-Chunk
                                        &     10.12                           &    54.20              & 48.96  
                \\
                                        & \algopt-Flatten                                             &       8.24                               &     43.91            &     36.48

                \\ \midrule    
                &\textit{Attention Speedup over \textcolor{blue}{the best decoding} }           & 1.50$\times$                                &  1.23$\times$                & 2.65$\times$

                \\
                \bottomrule
            \end{tabular}
        }

    \end{footnotesize}

\end{table}

\subsection{DeFT-Node Algorithm \label{A:DeFT-alg}}

\begin{algorithm}[!h]
    \caption{\algopt-Node Algorithm-Phase 1: QKV Preparation. }
    \label{alg:DeFT-alg_s1}
    \begin{algorithmic}
        \STATE {\bfseries Input:} query $Q \in R^{(b_q,d)}$, Key cache list $KL=(K_{0}, ...K_{N-1})$, Value cache list $VL=(V_{0}, ...V_{N-1})$ for each sequence node in the tree, where $N$  is the total number of sequences in a tree, and Tree $T$ with its topology information.
        \\

        \FOR{each $q$ in $Q$ with its global index $idx$}
        \STATE \textbf{/*Get KV indices of all prefixes' for a query.*/}
        \STATE $QMapKV[idx]$=GetPrefixKVIndices($q, KL, VL, T$)

        \ENDFOR

        \FOR{each seq's KV cache $K_{i},V_{i}$ in $KL, VL$ with its KV indice $i$}

        \STATE \textbf{/*Group each sequence's KV with all queries that share it.*/}
        \STATE $Q_i$= GroupQueryToKV($Q, K_{i}, V_{i}, T$) $\in R^{b_i,d} \subset Q$
        \STATE $KVMapQ[i]=Q_i $

        \ENDFOR
        \STATE \textbf{Return} QMapKV, KVMapQ

    \end{algorithmic}
\end{algorithm}

\algopt-Node has two phases-\textbf{Phase 1-QKV Preparation} and \textbf{Phase 2-Attention Calculation}.

\textbf{Phase 2-Attention Calculation} of \algopt has two stages.
\begin{enumerate}[nosep, leftmargin=12pt]
    \item \textbf{Stage 1: Calculate Partial Attentions.}  We will apply Flash Attention of all QKV groups obtained after \textbf{Phase 1-QKV Preparation} of \algopt, to get partial attention and LogSumExp.
    \item  \textbf{Stage 2: Global Reduction.}  We will remap  partial attention and LogSumExp based on each query, and get final attention based on global reduction similar to \flashdecoding\citep{dao2023flashdecoding}.
\end{enumerate}

\begin{algorithm}[!h]
    \caption{\algopt-Node Algorithm-Phase 2: Attention Calculation. }
    \label{alg:DeFT-alg}
    \begin{algorithmic}
        \STATE {\bfseries Input:} query $Q \in R^{(b_q,d)}$, Key cache list $KL=(K_{0}, ...K_{N-1})$, Value cache list $VL=(V_{0}, ...V_{N-1})$ for each sequence node in the tree, where $N$  is the total number of sequences in a tree, and Tree $T$ with its topology information.  QKV group information $QMapKV$, $KVMapQ$ from \textbf{QKV Preparation Phase}.
        \\

        \FOR{each $q$ in $Q$ with its global index $idx$}

        \STATE \textbf{/*Allocate to store LogSumExp of $Q@K^T$ grouped by query.*/}

        \STATE $LogSumExp[idx]=\{ \}$
        \STATE \textbf{/*Allocate to  store partial results of $SoftMax(Q@K^T)V$ for each query.*/}
        \STATE $O[idx]=\{ \}$

        \ENDFOR
        \STATE \textbf{/*Allocate space for output after reduction.*/}
        \STATE $FO= (0)_{b_q \times d} \in R^{(b_q,d)}$

        \FOR{each seq's KV cache $K_{i},V_{i}\in  R^{(b_{kv},d)},R^{(b_{kv},d)}$ in $KL, VL$ with its KV indice $i$}
        \STATE \textbf{\# Unroll for loop to SMs}

        \STATE $Q_i$= $KVMapQ[i] \in R^{(b_{i},d)}$
        \STATE \textbf{/*Get partial attention $o_i$ for each QKV group, LogSumExp $lse_i$ of $Q@K^T$ in row for reduction.*/}
        \STATE $o_i, lse_i$ = FlashAttention($Q_i,K_{i}, V_{i}$)
        \STATE $\in R^{(b_i,d)}, R^{b_i}$
        \STATE \textbf{/*Map the partial results back to each query for reduction.*/}
        \FOR{each query $q$ in $Q_i$ with its group index $gp\_idx$ and global index $idx$ in $Q$}
        \IF {$i \in QMapKV[idx]$}
        \STATE $LogSumExp[idx].append(lse_i[gp\_idx])$

        \ENDIF
        \ENDFOR
        \ENDFOR

        \FOR{each $q$ in $Q$ with its global index $idx$}
        \STATE \textbf{\# Unroll for loop to SMs}
        \IF{len($O[idx]$)==len($QMapKV[idx]$)}
        \STATE \textbf{/*Global reduction after collecting all partial results from QKV groups that contains $q$.*/}
        \STATE $LSE_{cat}$= CatTensor($LogSumExp[idx]$)
        \STATE $LSE_{max}$=RowMax($LSE_{cat}$)
        \STATE $Mid\_L=0 ,Mid\_O=0^ {(1,d)}$
        \FOR{each $lse_j$ in $LogSumExp[idx]$}
        \STATE $new\_exp=e^{lse_j-LSE_{max}}$
        \STATE $Mid\_L=Mid\_L+new\_exp$
        \ENDFOR
        \FOR{each $lse_j,o_j$ in $LogSumExp[idx],O[idx]$}
        \STATE $new\_exp=e^{lse_j-LSE_{max}}$
        \STATE $Mid\_O=Mid\_O+new\_exp@o_j/Mid\_L$
        \ENDFOR

        \STATE $FO[idx]=Mid\_O$
        \ENDIF
        \ENDFOR
        \STATE \textbf{Return $FO$}

    \end{algorithmic}
\end{algorithm}

\newpage
\subsection{\algopt-Flatten Algorithm \label{A:DeFT-Sub-alg}}
The algorithm (noted as \algopt-Node) in Appendix \ref{A:DeFT-alg} adopts a node-granularity split strategy, which is quite simple. However, when the token lengths of different nodes in a decoding tree are very unbalanced, it might introduce inefficient calculation due to the unbalanced workload in on-chip SMs of GPUs.

Therefore, we can split the decoding tree in a more balanced way-- in subtree-granularity. We show the \algopt-Flatten algorithm as follows, which also consists of two stages similar to \algopt-Node.

\begin{algorithm}[!h]
    \caption{\algopt-Flatten Algorithm-Phase 1: QKV Preparation. }
    \label{alg:DeFTS-alg_s1}
    \begin{algorithmic}
        \STATE {\bfseries Input:} query $Q \in R^{(b_q,d)}$, Key cache list $KL=(K_{0}, ...K_{N-1})$, Value cache list $VL=(V_{0}, ...V_{N-1})$ for each sequence node in the tree, where $N$  is the total number of sequences in a tree, and Tree $T$ with its topology information. Subtree size $S_t$, which means each subtree after tiling contains at most $S_t$ tokens.
        \\

        \STATE{\textbf{/*Evenly slice/blockwise the Tree \kvcache (with $n_T$ tokens in the tree ) to  subtrees.*/}}
        \STATE{SubInfo, KSub, VSub =Slice( KL, VL, $S_t$, $T$)}
        \STATE \textbf{/*Notes:
            (1) subtree number $m=Ceil(n_T/S_t)$; }
        \STATE \textbf{(2) subtrees' \kvcache $ KSub=(Kb_0,...,Kb_{m-1})$, $VSub=(Vb_0,...,Vb_{m-1})$;}
        \STATE \textbf{(3) subtree information $SubInfo=(Sb_0,...,Sb_{m-1})$, where each subtree i has $Sb_i=(ofs_0,...ofs_{n_{b_i}-1})$ to record the offset of each node in the subtree \kvcache, with $n_{b_i}$ as the total number of nodes in subtree $i$. */}

        \FOR{each subtree's KV cache $Kb_{i},Vb_{i}$ in $KSub, VSub$ with its subtree ID $i$}

        \STATE \textbf{/*Group each subtree's KV with all queries that share it.*/}
        \STATE $Q_i$= GroupQueryToKV($Q, Kb_{i}, Vb_{i}, T$) $\in R^{b_i,d} \subset Q$
        \STATE $KVMapQ[i]=Q_i $
        \FOR{each query $q$ in $Q_i$ with a global index $idx$ in $Q$}
        \STATE{$QMapKV[idx].append(i)$}
        \ENDFOR
        \STATE \textbf{/*Add a causal mask as different nodes in a subtree could be shared by different queries.*/}
        \STATE $CausalMask[i]=GetBitMask(Q_i,Kb_{i},Vb_{i},T)$=$(CM_0,...CM_{n_{b_i}-1})$
        \STATE where $n_{b_i}$ is the total number of nodes in the subtree, and $CM_i$ is a 64-bit int bit mask for node i.
        \STATE{\textbf{/*E.g, $100....00$ with 1 in bit 0, means the $Q_i[0]$ does not share \kvcache of node i in the subtree.*/}}
        \ENDFOR
        \STATE \textbf{Return} QMapKV, KVMapQ, CausalMask,SubInfo

    \end{algorithmic}
\end{algorithm}

\begin{algorithm}[!h]
    \caption{\algopt-Flatten Algorithm-Phase 2: Attention Calculation. }
    \label{alg:DeFT-alg_s2}
    \begin{algorithmic}
        \STATE {\bfseries Input:} query $Q \in R^{(b_q,d)}$, Key cache list in subtree-granularity KSub=($Kb_0$,...,$Kb_{m-1}$),  Value cache list in subtree VSub = ($Vb_0$,...,$Vb_{m-1}$ for $m$ subtrees after tiling based on Tree $T$ with its topology information.  QKV group information $QMapKV$, $KVMapQ$, causal mask $CausalMask$ and subtree information $SubInfo$  from \textbf{QKV Preparation Phase}.
        \\

        \FOR{each $q$ in $Q$ with its global index $idx$}

        \STATE \textbf{/*Allocate to store LogSumExp of $Q@K^T$ grouped by query.*/}

        \STATE $LogSumExp[idx]=\{ \}$
        \STATE \textbf{/*Allocate to  store partial results of $SoftMax(Q@K^T)V$ for each query.*/}
        \STATE $O[idx]=\{ \}$

        \ENDFOR
        \STATE \textbf{/*Allocate space for output after reduction.*/}
        \STATE $FO= (0)_{b_q \times d} \in R^{(b_q,d)}$

        \FOR{each subtree's KV cache $Kb_{i},Vb_{i} \in  R^{(b_{kv},d)},R^{(b_{kv},d)} $ in $KSub, VSub$ with subtree ID $i$}
        \STATE \textbf{\# Unroll for loop to SMs}
        \STATE $Q_i$= $KVMapQ[i] \in R^{(b_{i},d)}$
        \STATE \textbf{/*Reconstruct mask  for attention calculation based on $CausalMask$ and $SubInfo$*/}
        \STATE $bitmask=CausalMask[i]\in R^{n_{b_i}}$,where $n_{b_i}$ is the total number of nodes for subtree i.
        \STATE $SubOfst=SubInfo[i]\in R^{n_{b_i}}$
        \STATE $mask=ReconstructMask(bitmask,SubOfst) \in R^{(b_{i},b_{kv})}$
        \STATE \textbf{/*Get partial attention $o_i$ for each QKV group, LogSumExp $lse_i$ of $Q@K^T$ in row for reduction.*/}
        \STATE $o_i, lse_i$ = FlashAttention($Q_i,Kb_{i}, Vb_{i},mask$)
        \STATE $\in R^{(b_i,d)}, R^{b_i}$
        \STATE \textbf{/*Map the partial results back to each query for reduction.*/}
        \FOR{each query $q$ in $Q_i$ with its group index $gp\_idx$ and global index $idx$ in $Q$}
        \IF {$i \in QMapKV[idx]$}
        \STATE $LogSumExp[idx].append(lse_i[gp\_idx])$

        \ENDIF
        \ENDFOR
        \ENDFOR

        \FOR{each $q$ in $Q$ with its global index $idx$}
        \STATE \textbf{\# Unroll for loop to SMs}
        \IF{len($O[idx]$)==len($QMapKV[idx]$)}
        \STATE \textbf{/*Global reduction after collecting all partial results from QKV groups that contains $q$.*/}
        \STATE $LSE_{cat}$= CatTensor($LogSumExp[idx]$)
        \STATE $LSE_{max}$=RowMax($LSE_{cat}$)
        \STATE $Mid\_L=0 ,Mid\_O=0^ {(1,d)}$
        \FOR{each $lse_j$ in $LogSumExp[idx]$}
        \STATE $new\_exp=e^{lse_j-LSE_{max}}$
        \STATE $Mid\_L=Mid\_L+new\_exp$
        \ENDFOR
        \FOR{each $lse_j,o_j$ in $LogSumExp[idx],O[idx]$}
        \STATE $new\_exp=e^{lse_j-LSE_{max}}$
        \STATE $Mid\_O=Mid\_O+new\_exp@o_j/Mid\_L$
        \ENDFOR

        \STATE $FO[idx]=Mid\_O$
        \ENDIF
        \ENDFOR
        \STATE \textbf{Return $FO$}

    \end{algorithmic}
\end{algorithm}

\end{document}